\documentclass[sn-mathphys,iicol,Numbered]{sn-jnl}

\usepackage{graphicx}%
\usepackage{amsmath,amsfonts,amssymb,amsthm, bm, bbm}

\usepackage{mathrsfs}%
\usepackage[title]{appendix}%
\usepackage[dvipsnames]{xcolor}%
\usepackage{textcomp}%
\usepackage{manyfoot}%
\usepackage{booktabs}%
\usepackage{algorithm}%
\usepackage{algorithmicx}%
\usepackage{algpseudocode}%
\usepackage{listings}%
\usepackage{cuted}
\usepackage{multicol}

\usepackage{comment}
\usepackage{tikz}
\usepackage{dcolumn}

\usepackage{lineno}

\newcommand\underrel[2]{\mathrel{\mathop{#2}\limits_{#1}}}

\tikzstyle{int}=[draw, fill=blue!20, minimum size=2em]
\tikzstyle{init} = [pin edge={to-,thin,black}]
\usetikzlibrary{shapes.misc, shapes, positioning,fit,calc}

\usepackage{makecell}

\usetikzlibrary{arrows,shapes,calc,trees,positioning,chains,shapes.geometric,decorations.pathreplacing,decorations.pathmorphing, matrix,shapes.symbols}
\usepackage{varwidth}

\usepackage{subcaption}

\tikzset{
	>=stealth',
	box/.style={
		rectangle,
		rounded corners,
		dashed,
		draw=black, very thick,
		minimum height=2em,
		text centered,
		execute at begin node={\begin{varwidth}{28em}},
		execute at end node={\end{varwidth}}},
	solidbox/.style={
		rectangle,
		rounded corners,
		draw=black, very thick,
		minimum height=2em,
		text centered,
		execute at begin node={\begin{varwidth}{28em}},
			execute at end node={\end{varwidth}}},
	fw_arrow/.style={
		->,
		thick,
		shorten <=2pt,
		shorten >=2pt,},
	bw_arrow/.style={
		<-,
		thick,
		shorten <=2pt,
		shorten >=2pt,},
	solidbox2/.style={
		rectangle,
		rounded corners,
		draw=black, very thick,
		minimum height=2em,
		text centered}
}

\DeclareMathOperator*{\argmin}{arg\,min}

\allowdisplaybreaks

\title[title]{Unifying supervised learning and VAEs - coverage, systematics and goodness-of-fit in normalizing-flow based neural network models for astro-particle reconstructions}

\author*[1]{Thorsten Gl\"usenkamp}\email{thorsten.gluesenkamp@physics.uu.se}

\affil[1]{\orgdiv{Department of physics}, \orgname{Uppsala University}, \orgaddress{\city{Uppsala}, \country{Sweden}}}

\begin{document}

\abstract{Neural-network based predictions of event properties in astro-particle physics are getting more and more common. However, in many cases the result is just utilized as a point prediction. Statistical uncertainties, coverage, systematic uncertainties or a goodness-of-fit measure are often not calculated. Here we describe a certain choice of training and network architecture that allows to incorporate all these properties into a single network model.
We show that a KL-divergence objective of the joint distribution of data and labels allows to unify supervised learning and variational autoencoders (VAEs) under one umbrella of stochastic variational inference.  
The unification motivates an \textit{extended} supervised learning scheme which allows to calculate a goodness-of-fit  p-value for the neural network model. Conditional normalizing flows amortized with a neural network are crucial in this construction. We discuss how to calculate coverage probabilities without numerical integration for specific "base-ordered" contours that are unique to normalizing flows. Furthermore we show how systematic uncertainties can be included via effective marginalization during training. The proposed extended supervised training incorporates (1) coverage calculation, (2) systematics and (3) a goodness-of-fit measure in a single machine-learning model. There are in principle no constraints on the shape of the involved distributions, in fact the machinery works with complex multi-modal distributions defined on product spaces like $\mathbb{R}^n \times \mathbb{S}^m$. The coverage calculation, however, requires care in its interpretation when the distributions are too degenerate. We see great potential for exploiting this per-event information in event selections or for fast astronomical alerts which require uncertainty guarantees.
}

\maketitle


\section{Introduction}
\label{sec:introduction}

Deep neural networks have achieved great results over the last couple of years. While the breakthroughs were initially made in industrial applications, for example in image processing \cite{Krizhevsky2012}, in recent years their application in fundamental science has become more and more ubiquitous. 
A typical application of deep learning in experimental high-energy physics is concerned with the reconstruction of particle interactions \cite{guest_review_high_energy}. These include discrete quantities like particle type \cite{lhc_classification} or continuous quantities like direction, position and energy \cite{murach2015_nn}. Traditionally, reconstructions of continuous quantities are performed by parametrized likelihood fits \cite{amanda_track_reco} which allow to calculate confidence intervals with standard Frequentist or Bayesian methods \cite{Roe2012}. Neural networks on the contrary are often used to produce point estimates \cite{nn_aartsen_2019}, and there is no universal agreed-upon notion how to calculate confidence intervals. Often they are just ignored, the result is registered as an observable and used in a binned likelihood analysis \cite{murach2015_nn}. Per-event uncertainties are not necessarily required for this use-case. However, there are many situations where precise uncertainty quantification is important. An example are per-event reconstructions in high-energy neutrino telescopes like IceCube \cite{Aartsen2017} or gravitational-wave observatories like LIGO \cite{Abbott2009} in the  context of multi-messenger astronomy \cite{icrc_multimessenger}. These experiments send out alerts to the astronomical community to perform follow-up observations with other experiments. On the one hand, these alerts are time-critical and should be sent out to the public as fast as possible. On the other hand, the uncertainties of the inferred direction must be precise and in particular not biased. While the time-critical aspect of this use-case is in favor of neural networks, obtaining unbiased uncertainties can be a challenge. A naive solution could employ Bayesian neural networks \cite{bayesian_nns} or approximations like certain dropout-variations \cite{mc_dropout} or ensemble-methods \cite{ensemble_methods}. However, these methods model a posterior over network weights, not over the actual physics parameters. An alternative is to parametrize the likelihood function with a neural network and perform a standard likelihood-based Frequentist or Bayesian analysis. Recently, this was done for gravitational-wave signals with flow-based networks \cite{Wong2020}. Such methods, however, inherit the disadvantages of likelihood-based inference. They can be fall into local optima during an optimization routine or have potentially long running times when Markov Chain Monte Carlo \cite{mcmc} (MCMC) is used to obtain samples. An alternative that recently has gotten popular is likelihood-free inference with neural networks, in which the posterior is directly learned from data. This has been applied to gravitational wave posteriors modeling the posterior as a parametrized Gaussian or mixture of Gaussians \cite{gaussian_posterior_modelling} and going beyond to use autoregressive normalizing flows for more complex shapes \cite{Green2020}. This paper is about the same line of thinking, where the posterior or more generally parts of the joint distribution of data and labels are learnt using normalizing flows.

We first discuss that the training process in supervised learning can be recast as variational inference of the true posterior distribution over labels, where the variational approximation is parametrized by a neural network. The derivation involves the Kullback-Leiber (KL)-divergence \cite{Kullback1951a} of the joint distribution of data and labels. This is a known derivation and the final loss function is sometimes called "conditional maximum-likelihood objective" in the machine learning literature \cite{Martens2014}. While the loss indeed represents a likelihood with respect to the neural network parameters, we emphasize it is more useful to think of it as an approximation of the posterior over the latent variables. Standard supervised learning usually uses the mean-squared-error (MSE) loss function, which corresponds to a standard-normal posterior as an approximation of the true posterior. This can often be a bad approximation. We compare it to the slightly more complex case of a Gaussian with a single covariance parameter and in particular approximations based on normalizing flows \cite{nf_review}, which in principle have arbitrary approximation capabilities. Since all Gaussian approximations, including the basic MSE loss, can be thought of as special cases of normalizing flows, so-called \emph{affine flows}, it seems sensible to view all supervised loss functions from this angle. While normalizing-flow base distributions can be arbitrary, it turns out that there are advantages of starting with a standard normal as a base distribution that go beyond the argument of simple evaluation. The standard normal automatically allows for straight-forward coverage tests, which are discussed in Section \ref{sec:coverage}.

Usually, variational inference with neural networks is employed in unsupervised learning in the context of variational autoencoders (VAEs). As we will show (Section \ref{sec:unsupervised}), one can derive the Evidence Lower Bound (ELBO) \cite{variational_inference_intro} of a variational autoencoder from the same joint KL-divergence as the supervised "maximum likelihood objective". This derivation not only explicitly indicates how supervised learning and VAE training is connected, but it also sheds some light on the interpretation of VAEs. Importantly, we use this connection to motivate a mixed training scheme which we call "extended supervised" learning (Section \ref{sec:extended_supervised}). This allows to calculate Bayesian goodness-of-fit values (Section \ref{sec:goodness_of_fit}) for the trained model. Figure \ref{fig:overview} indicates an overview of the resulting picture. Traditional variational inference (Fig. \ref{fig:overview} a) is typically discussed on a per-event level or in the context of neural networks as approximating the posterior over the network parameters in Bayesian neural networks \cite{bayesian_nns}. However, one is really interested in the posterior over physics parameters, ideally for all events simultaneously. Figure \ref{fig:overview} b) illustrates that supervised learning, extended supervised learning and unsupervised VAEs can be interpreted as stochastic variational inference using the KL-divergence of the joint distribution of data, observed and unobserved labels (latent variables). Because only sub-parts of the joint distribution are effectively parametrized in these approaches, one obtains "explicit" variational inference solutions for the posterior distributions, in the sense that the conditional structure of the posterior is explicitly modeled and thereby different for each datapoint. This is sometimes also called "local" variational inference \cite{variational_inference_intro} in the literature.
The unified viewpoint in combination with amortized conditional normalizing flows naturally leads to answers to the following questions:

\begin{enumerate}
	\item How can we do coverage tests with neural networks on complex base distributions, including distributions over directions? 
	\item How are systematic uncertainties incorporated in the training process?
	\item How can we do goodness-of-fit checks on neural-network predictions?
\end{enumerate}

The first half of the paper is concerned with the unified viewpoint and derives the various loss functions from the joint KL-divergence. The second half then answers the three questions above.

\begin{figure*}
    \centering
    \begin{subfigure}[t]{0.99\textwidth}
        \centering
        \hspace*{-1,05cm}%
        \begin{tikzpicture}[]

\node (normal_single) [solidbox2, draw,inner sep=10pt, rounded corners, minimum size=2cm,anchor=east] {
	
	\begin{tabular}{c}
	\\
    \textbf{Input:} datum $x_j$, 
	        likelihood $\mathcal{P}_t(x; z)$ \\
     \\VI posterior (via ELBO):
     \\
     $q_{\lambda}(z) \approx \mathcal{P}_t(z;x_j)$
     \\
    \multicolumn{1}{l}{\textbf{Issue:}}
     \\
     only a single event, requires likelihood
    \end{tabular}

	};
	\node[solidbox2, fill=blue!20] at (normal_single.north) {Single event};
	   
	  \node (normal_nn) [solidbox2, draw,inner sep=10pt, rounded corners, minimum size=2cm, right=5pt] at (normal_single.east) {
	
	\begin{tabular}{c}
	\\
    \textbf{Input:} data $x_i$, labels $z_i$ ($i=1\ldots N$)\\
     \\VI posterior (via ELBO):
     \\
     $q_{\lambda}(w) \approx \mathcal{P}_t(w)$
     \\
     \multicolumn{1}{l}{\textbf{Issue:}}
     \\
     interested in posterior over $z$, not over $w$
    \end{tabular}

	};
	\node[solidbox2, fill=blue!20] at (normal_nn.north) {VI of NN weights};

\end{tikzpicture}
        \caption{Two examples of "classical" variational inference. The variational parameters are denoted by $\lambda$.}
    \end{subfigure}%
    
    \begin{subfigure}[t]{0.99\textwidth}
        \centering
        \hspace*{-1,3cm}%
        \begin{tikzpicture}[]
\node(shadow) [] {};
\node (sv) [solidbox2, draw,inner sep=10pt, rounded corners, minimum size=2cm, below of = shadow, yshift=-2.5cm] {
	
	\begin{tabular}{c}
	\\
    \textbf{Input:} data $x_i$, labels $z_{o,i}$ \\
    \\
    $D_{\mathrm{KL}} (\mathcal{P}_t|q)[x,z_o]$
    \\ \\
     \\VI posterior \\ (via supervised loss):
     \\
     $q_{\phi}(z_o;x) \approx \mathcal{P}_t(z_o;x)$
 
    \end{tabular}
	};
	
	\node(title_sv) [solidbox2, fill=green!20] at (sv.north) {supervised learning};
	
    \node (phantom1_sv) [below of=title_sv,yshift=-1.0cm] {};
    \node (phantom2_sv) [below of=title_sv,yshift=-2.0cm] {};
    \path[->] (phantom1_sv) edge (phantom2_sv);
    
    
    \node (ext) [solidbox2, draw,inner sep=10pt, rounded corners, minimum size=2cm, right=5pt] at (sv.east) {
	
	\begin{tabular}{c}
	\\
    \textbf{Input:} data $x_i$, labels $z_{o,i}$  \\
    \\
    $D_{\mathrm{KL}} (\mathcal{P}_t|q)[x,z_o,z_u]$
    \\ \\
     \\VI posterior \\ (via mixed loss):
     \\
     $q_{\phi,\varphi}(z_o,z_u;x) \sim \mathcal{P}_t(z_o,z_u;x)$
 
    \end{tabular}
	};

	\node (title_ext)[solidbox2, fill=green!20] at (ext.north) {ext. supervised learning};
	\node (phantom1_ext) [below of=title_ext,yshift=-1.0cm] {};
    \node (phantom2_ext) [below of=title_ext,yshift=-2.0cm] {};
    \path[->] (phantom1_ext) edge (phantom2_ext);
    
    
    \node (vae) [solidbox2, draw,inner sep=10pt, rounded corners, minimum size=2cm, right=5pt] at (ext.east) {
	
	\begin{tabular}{c}
	\\
    \textbf{Input:} data $x_i$ \\
    \\
    $D_{\mathrm{KL}} (\mathcal{P}_t|q)[x,z_u]$
    \\ \\
     \\VI posterior \\ 
     (via ELBO):
     \\
     $q_{\phi}(z_u;x) \sim \mathcal{P}_t(z_u;x)$
 
    \end{tabular}
	};
	
	\node (title_vae) [solidbox2, fill=green!20] at (vae.north) {VAE learning};
	   
	 \node (phantom1_vae) [below of=title_vae,yshift=-1.0cm] {};
    \node (phantom2_vae) [below of=title_vae,yshift=-2.0cm] {};
    \path[->] (phantom1_vae) edge (phantom2_vae);
\end{tikzpicture}
        \caption{Three examples of "explicit" variational inference of the posterior in target space $z$, unified by the joint KL-divergence as outlined in Section \ref{sec:joint_kl_divergence_derivations}. The label ($z_o$ - "observed") and latent ($z_u$ - "unobserved") variables correspond to physical parameters in high-energy physics. For the latent variables, this is only true up to an identifiability class, denoted by "$\sim$" instead of "$\approx$" (see Section \ref{sec:joint_kl_divergence_derivations} for more details). The variational parameters are denoted by $\phi$ and $\varphi$ to match the nomenclature in the rest of the paper. }
    \end{subfigure}
    
    \caption{Variational inference (VI) examples for simulation data $x_i$ and labels $z_i$ indexed by $i=1 \ldots N$, which comprise the whole dataset of size $N$. The exception is the single event example in (a), which has a single datum $x_j$ as input. }
   \label{fig:overview}
\end{figure*}

\section{Monte Carlo estimates and the joint KL-divergence}
\label{sec:joint_kl_divergence_derivations}
Physics experiments perform measurements on the final outcomes of a causal chain of events.
These measurements are inherently probabilistic due to noise and the randomness from particle-physics interactions, and the measured observable follows a specific probability distribution. The probability distribution over possible measurement outcomes is called the likelihood function when viewed as a function of its parameters. It is a central object in both Frequentist and Bayesian statistical analyses to perform parameter estimation \cite{Roe2012}. Its shape is entirely determined by the laws of nature in combination with the detector response. However, because the laws can be convoluted and the experiment can be very complex, it is usually not possible to write down an explicit analytic expression. A common practice is to estimate the likelihood function from Monte Carlo simulations where all these complex effects are considered \cite{bsplines}. This estimated likelihood function is then used to perform inference or calculate confidence intervals.

Neural networks allow to skip this estimation step completely, because the Monte Carlo samples themselves are drawn from the true joint distribution $\mathcal{P}_t(x,z_o)$ of observations $x$ and parameters of interest $z_o$. 
Let us call the corresponding true data generating function $\mathcal{P}_t(x;z_o)$ which is the true probability distribution of the measured data $x$ given the properties $z_o$.
Here $z_o$ stands for recorded or \textbf{\emph{o}}bserved properties in the simulation, for example the position of a particle interaction. Connected to this data generating function, there exists a true posterior distribution $\mathcal{P}_t(z_o;x)$ and a true prior distribution $\mathcal{P}_t(z_o)$. The true joint distribution follows the distribution which includes the detector response and selection effects inherent to the measurement - it also includes artificial selection effects. For example, if the generating function of the direction of injected particles is uniform, the actually recorded Monte Carlo events will generally not be uniform due to detector effects. The implicitly contained true prior $\mathcal{P}_t(z_o)$ is this non-uniform distribution over $z_o$ of the actually registered events, not the uniform generating distribution.

Since a Monte Carlo simulation draws samples $x_i,z_{o,i}$ from the joint distribution $\mathcal{P}_t(x,z_o)$, we can always evaluate any expectation value under the true joint probability distribution as 
\begin{align}
E_{x,z_o}[f(x,z_o)] &= \int\limits_{x,z_o} \mathcal{P}_t(x,z_o) f(x,z_o) \ dx \ dz_o \\
&\approx \frac{1}{N} \sum\limits_{{x_i,z_{o,i}}} f(x_i, z_{o,i}) ,
\end{align} where $i$ indexes the $N$ samples. The following sections make use of a specific choice for $f$, namely $f=\mathrm{ln}\frac{\mathcal{P}_t(x,z_o)}{q(x,z_o)}$, which yields an expectation value that equals the KL-divergence between two distributions $\mathcal{P}_t(x,z_o)$ and $q(x,z_o)$. The KL-divergence \cite{Kullback1951a} is a natural quantity to measure the distance between two probability distributions. In particular, if $q_\phi$ is a parametrized probability distribution, the KL-divergence defines a loss function over $\phi$ that achieves its minimum when $q_\phi$ is equal to $\mathcal{P}_t$. It is therefore often used in \textit{variational} methods which perform inference via optimization \cite{variational_inference_intro}. It turns out that the joint KL-divergence can be used to derive the loss functions in both supervised learning and unsupervised VAEs and thereby unifies them as two connected approaches to variational inference in slightly different circumstances.

\subsection{Supervised learning}
\label{sec:supervised}

The KL-divergence of the true joint distribution $\mathcal{P}_t(z_o,x)$ and an approximation $q_\phi(z_o,x)$ can be written as

\begin{align}
&\ \ \ D_{\mathrm{KL, joint(x,z_o)}} (\mathcal{P}_t;q_\phi) \nonumber \\
&=\int\limits_{x}^{}\int\limits_{z_o}^{} \mathcal{P}_t(z_o,x) \cdot \mathrm{ln} \frac{\mathcal{P}_t(z_o,x)}{q_\phi(z_o,x)} dz_o dx \\
\begin{split}
&= \int\limits_{x}^{}\int\limits_{z_o}^{} \mathcal{P}_t(z_o;x)\cdot \mathcal{P}_t(x) \\
&\ \ \ \cdot \mathrm{ln} \frac{\mathcal{P}_t(z_o;x)\cdot \mathcal{P}_t(x)}{q_\phi(z_o;x) \cdot q(x)} dz_o dx 
\end{split}
\\
\begin{split}
&= \int\limits_{x}^{}\int\limits_{z_o}^{} \mathcal{P}_t(z_o;x)\cdot \mathcal{P}_t(x) \\
&\ \ \ \cdot \left( \mathrm{ln} \frac{\mathcal{P}_t(z_o;x)}{q_\phi(z_o;x)} + \mathrm{ln}\frac{\mathcal{P}_t(x)}{q(x)} \right) dz_o dx
\end{split}
\\
\begin{split}
&=\int\limits_{x}^{}\int\limits_{z_o}^{} \mathcal{P}_t(x) \cdot \mathcal{P}_t(z_o;x) \cdot  \mathrm{ln} \frac{\mathcal{P}_t(z_o;x)}{q_\phi(z_o;x)} dz_o dx \\
&\ \ \ + \int\limits_{x}^{} \mathcal{P}_t(x) \cdot  \mathrm{ln}\frac{\mathcal{P}_t(x)}{q(x)}  dx 
\end{split}
\\
\begin{split}
&= E_{x} [D_{\mathrm{KL}}(\mathcal{P}_t(z_o;x);q_\phi(z_o;x)] \\ 
&\ \ \ + D_{\mathrm{KL}}(\mathcal{P}_t(x);q(x)).\label{eq:supervised_loss}
\end{split}
\end{align}
The distributions involving $\mathcal{P}_t$ can not be evaluated analytically, but as discussed above Monte Carlo simulations yield samples from $\mathcal{P}_t$ and hence provide sample-based estimates of the integrals. We only parametrize the conditional distribution over labels, $q_\phi(z_o;x)$ with $\phi$, and leave the distribution $q(x)$ unparametrized as it is typically not of interest. Taking the sample estimate yields the following update rule over $\phi$ to minimize the KL-divergence objective and thereby minimize the supervised loss function $\mathcal{L}_{\mathrm{supervised}}(\phi)$:

\begin{strip}
\begin{align}
 \argmin_{\phi} \hat{D}_{\mathrm{KL, joint(x,z_o)}} (\mathcal{P}_t;q_{\phi}) &= \argmin_{\phi} \frac{1}{N} \sum_{ {x_i,z_{o,i}}} \mathrm{ln} \left (\frac{ \mathcal{P}_t(z_{o,i};x_i)}{ q_{\phi}(z_{o,i};x_i)}\right)   +\mathrm{ln} \left (\frac{ \mathcal{P}_t(x_i)}{ q(x_i)}\right)  \\
&= \argmin_{\phi}  \frac{1}{N} \sum\limits_{ {x_i,z_{o,i}}}^{} - \mathrm{ln}  \left( q_{\phi}(z_{o,i};x_i)\right) + \mathrm{const} \label{eq:real_supervised_loss} \\
&= \argmin_{\phi} \frac{1}{N}  \sum\limits_{{x_i,z_{o,i}}}^{} - \mathrm{ln}  \left( q_{\phi}(z_{o,i};x_i)\right) \\ &\equiv \argmin_{\phi} \mathcal{L}_{\mathrm{supervised}}(\phi).
\end{align}
\end{strip}
The approximation of the marginal likelihood, $q(x_i)$, can be dropped as a constant part with respect to changes in $\phi$. Additionally we can drop the true posterior evaluations $\mathcal{P}_t(z_{o,i};x_i)$ which are also constant with respect to $\phi$.
If the distribution $q_\phi$ is actually parametrized by a network whose parameters are $\phi$, the derivation shows that minimizing the KL-divergence between the true posterior and an approximation given by a neural network is equivalent to standard neural network training where the goal is to minimize negative log-probability over labels. This has been discussed previously \cite{Martens2014}, but it is usually described as a maximum likelihood objective with respect to the network parameters. Here, we emphasize that it is really more useful to think of $q_{\phi}(z_{o};x_i)$ as a parametrized posterior over labels $z_{o}$ given the data $x_i$, not as a conditional likelihood function with respect to the parameters $\phi$.  
In unsupervised learning the labels $z_{o,i}$ are not available so the precise above objective does not work.  
In the following we show that a simple replacement of the posterior KL-divergence in eq. \ref{eq:supervised_loss} with the respective reverse KL-divergence leads to a tractable solution via the reparametrization trick \cite{kingma2013} which contains the evidence lower bound (ELBO) and thereby the VAE  objective. 

\subsection{Unsupervised learning: Variational autoencoders}
\label{sec:unsupervised}

The following derivation is motivated by an experimental physicists' point of view when there is access to a Monte Carlo simulation. Here, one way to think of latent variables in unsupervised variational autoencoders is to imagine them as unobserved labels that are not recorded in the simulation. The implied direct comparison, and renaming, of "observed" into "unobserved" labels is justified if the latent variables have positive mutual information with the data. Mutual information is a non-linear generalization of correlation \cite{mutual_information}. In events of a static particle physics detector with a single particle interaction, for example, the only properties that can have positive mutual information with the data are properties of the particle interaction. These properties can be in principle be labeled. 
 Typical labels are the position or direction of a particle at the interaction point. \footnote{If the number of latent variables is very large, some of them will be superfluous and not carry any information about the data. For such variables this identification strictly makes no sense since they could never be labeled.}

The aim of the following exercise is to start with the same KL-divergence as for the supervised loss, but with unobserved labels $z_u$ instead of observed labels $z_o$, and deduce which modifications have to be made in order to obtain the ELBO loss of a variational autoencoder as a partial tractable objective. It is not the aim to deduce strict new numerical results, but to indicate how supervised learning and unsupervised learning are precisely connected in the variational viewpoint.
 
 Starting with the same joint KL-divergence, but now using unobserved labels $z_u$, we can write 
 
\begin{align}
\begin{split}
&\ \ \ \ D_{\mathrm{KL, joint(x,z_u)}} (\mathcal{P}_t;q_\phi) \\ &= E_{x} \left[ \int\limits_{z_u}^{} \mathcal{P}_t(z_u;x) \cdot  \mathrm{ln} \frac{\mathcal{P}_t(z_u;x) \cdot \mathcal{P}_t(x)}{q_\phi(z_u;x) \cdot q(x)} dz_u \right] 
\end{split}
\\
\begin{split}
&=E_{x} [\textcolor{orange}{D_{\mathrm{KL}}(\mathcal{P}_t(z_u;x);q_\phi(z_u;x)}] \\
&\ \ \ + \underbrace{D_{\mathrm{KL,Marg(x)}}(\mathcal{P}_t(x);q(x))}_{\equiv D_{\mathrm{KL,M, const}}},
\end{split}
\end{align}
which results in two intractable terms.  The second term $D_{\mathrm{KL,M, const}}$ can for further discussion be ignored since $q(x)$ is typically not parametrized and therefore irrelevant in optimization schemes. In the first term, the outer integral over $x$ is explicitly kept, because we do have samples from $x$ and so we can in principle evaluate expectation values over $x$. The inner part, however, involves an expectation over the intractable KL-divergence of the conditional distribution $\mathcal{P}_t(z_u;x)$ with $q_\phi(z_u;x)$ (\textcolor{orange}{orange}). In order to proceed to some extent, we could replace the KL-divergence  within the expectation value with any generalized divergence measure $\mathcal{D}(\mathcal{P}_t;q)$, for example f-divergences \cite{f_divergences}, as long as it shares the property that it has a minimum when $\mathcal{P}_t(z_u;x)=q_\phi(z_u;x)$. However, there is only a subset that will straightforwardly lead to the ELBO loss of the VAE. Among those, a natural choice is the reverse KL-divergence, $D_{\mathrm{rev. KL}}(q_\phi(z_u;x);\mathcal{P}_t(z_u;x))$, which we choose for simplicity. This leads to the following surrogate loss term:

\begin{align}
\begin{split}
&\ \ \ \ \mathrm{Surrogate}_{\mathrm{KL}}(\mathcal{P}_t;q_{\phi}) \\
&= E_{x} [\textcolor{NavyBlue}{D_{\mathrm{rev. KL}}(q_{\phi}(z_u;x);\mathcal{P}_t(z_u;x)}] \\
&\ \ \ + \underbrace{D_{\mathrm{KL,Marg(x)}}(\mathcal{P}_t(x);q(x))}_{\equiv D_{\mathrm{KL,M, const}}}.
\end{split}
\end{align}

As long as $q_\phi$ has arbitrary approximation capabilities, the surrogate term has the property that the minimum over the parameters $\phi$ is equal to the minimum of the original KL-divergence

\begin{align}
\begin{split}
&\argmin_{\phi} {D_\mathrm{KL, joint(x,z_o)}} (\mathcal{P}_t;q_{\phi}) \\
= &\argmin_{\phi} \mathrm{Surrogate}_{\mathrm{KL}}(\mathcal{P}_t;q_{\phi})  \label{eq:reverse_switch},
\end{split}
\end{align}
because both KL-divergence (\textcolor{orange}{orange}) and reverse KL-divergence (\textcolor{NavyBlue}{blue}) are equal to zero when the two involved distributions are equal. So far this is just a theoretical exercise, as $\mathcal{P}_t$ is inaccessible. The surrogate loss, however, contains the ELBO loss.
In order to see this, we can rewrite the surrogate loss as

\begin{align}
\begin{split}
 &\ \ \ \ \mathrm{Surrogate}_{\mathrm{KL}}(\mathcal{P}_t;q_\phi) \\
 &= E_{x} [D_{\mathrm{rev. KL}}(q_\phi(z_u;x);\mathcal{P}_t(z_u;x))] \\
 &\ \ \ + D_{\mathrm{KL,M}}  
\end{split}
\\
&= E_{x} \left[ \int\limits_{z_u}^{} q_\phi(z_u;x) \cdot  \mathrm{ln} \frac{q_\phi(z_u;x) }{\mathcal{P}_t(z_u;x)} dz_u \right] + D_{\mathrm{KL,M}}\\
\begin{split}
&= \underbrace{E_{x} \left[ \int\limits_{z_u}^{} q_\phi(z_u;x) \cdot  \mathrm{ln} \ q_\phi(z_u;x) dz_u \right]}_{\mathrm{neg. \ entropy \ term}} \\ 
&\ \ \ + E_{x} \left[ \int\limits_{z_u}^{} q_\phi(z_u;x) \cdot  \mathrm{ln} \frac{\mathcal{P}_t( x)}{\mathcal{P}_t(x,z_u)} dz_u \right] + D_{\mathrm{KL,M}} \label{eq:pure_entropy} 
\end{split}
\\
\begin{split}
&= \underbrace{E_{x} \left[ \int\limits_{z_u}^{} q_\phi(z_u;x) \cdot  \mathrm{ln} \frac{q_\phi(z_u;x) \cdot \mathcal{P}_t(x)}{p_\theta(x;z_u)\cdot p_\psi(z_u)} dz_u \right]}_{D_{KL}(\mathcal{P}_t(x)\cdot q_\phi(z_u;x), p_{\theta/\psi}(x,z_u))} \\ 
&\ \ \ + \underbrace{E_{x} \left[ \int\limits_{z_u}^{} q_\phi(z_u;x) \cdot  \mathrm{ln} \frac{p_\theta(x;z_u) \cdot p_\psi(z_u)}{\mathcal{P}_t(x,z_u)} dz_u \right]}_{\equiv \mathcal{R}(\phi, \theta, \psi)} + D_{\mathrm{KL,M}} \label{eq:r_introduction}
\end{split}
\\
\begin{split}
&= E_x [\mathrm{ln} \mathcal{P}_t(x)] \\
&- E_x \underbrace{\left[ \int\limits_{z_u}^{} q_\phi(z_u;x) \cdot \left[  \mathrm{ln} p_\theta(x;z_u) - \mathrm{ln}\frac{q_\phi(z_u;x)}{p_\psi(z_u)}  \right] dz_u \right]}_{\mathrm{evidence \ lower \   bound \ (ELBO)}} \\ 
&\ \ \ + \mathcal{R}(\phi, \theta, \psi) \\
&\ \ \ +  D_{\mathrm{KL,M}} ,
\end{split}
\end{align}
which results in the negative ELBO loss, two constant terms, and a residual term $\mathcal{R}(\phi, \theta, \psi)$. 
The important step in this reformulation is the introduction of an auxiliary distribution $p_{\theta/\psi}(z_u,x)=p_{\theta}(x;z_u)\cdot p_{\psi}(z_u)$ from eq. \ref{eq:pure_entropy} to eq. \ref{eq:r_introduction}. The negative entropy term in eq. \ref{eq:pure_entropy} by itself is tractable, but behaves divergent when minimized over $\phi$, since the negative entropy will tend to infinity. Therefore, the introduction of $p_{\theta/\psi}(z_u,x)$ is a necessary ingredient to obtain anything that has a chance for non-trivial behavior. We also indicate by the subscript $\psi$ that the prior $p_\psi(z_u)$ can be parametrized by $\psi$, even though in many applications it is just taken to be a fixed standard normal distribution. 
With the introduction of $p_{\theta/\psi}(x,z_u)$, the surrogate KL-divergence then splits into two terms in eq. \ref{eq:r_introduction}. 

The first term is the KL-divergence between $\mathcal{P}_t(x)\cdot q(z_u;x)$ and $p_{\theta/\psi}(x,z_u)$, which is equal to the ELBO loss when the constant term $E_x [\mathrm{ln} \mathcal{P}_t(x)]$ is pulled out of the integral. This joint KL-divergence between $\mathcal{P}_t(x)\cdot q_\phi(z_u;x)$ and $p_{\theta/\psi}(x,z_u)$ has been used before in the context of the InfoVAE \cite{Zhao2017} or more general VAE architectures with additional constraints \cite{Zhao2018} and is by itself another non-standard starting point to derive the ELBO loss. Here it arises as a product in the derivation which is connected to the supervised loss derivation via the KL-divergence of the joint distribution. It is important to note that we have to use this more complicated construction, compared to just start with this simpler KL divergence, in order to see the similarity to the supervised loss derivation and then be equipped with a canonical way to derive the extended supervised case in Section \ref{sec:extended_supervised}.

The second term is a residual term $\mathcal{R}(\phi, \theta, \psi)$. This term is not tractable, because any samples drawn from $q_\phi(z_u;x)$ can not be evaluated by the inaccessible density $\mathcal{P}_t(z_u;x)$. However,
one can deduce that after an ELBO optimization with solution $\theta^*$ and $\phi^*$ and flexible enough density parametrizations, the residual term is always bounded from below by zero because it is equal to a proper KL-divergence:

\begin{align}
&\ \ \ \ q_{\phi^*}(z_u;x)\cdot \mathcal{P}_t(x) \approx p_{\theta^*/\psi^*}(x,z_u) \\
\begin{split}
&\rightarrow \mathcal{R}(\phi^*, \theta^*, \psi^*) \\
&\ \ \ \ \approx D_{\mathrm{KL}}(q_{\phi^*}(z_u;x)\cdot\mathcal{P}_{t}(x), \mathcal{P}_{t}(z_u,x)) \geq 0.
\end{split}
\end{align}
This construction therefore makes it explicit that ELBO optimization alone does not necessarily lead to a joint density that matches the true density because $\mathcal{R}$ can be greater than zero - a fact that is lost in the standard ELBO derivation \cite{jordan_elbo_derivation} based on the marginal likelihood and Jensen's inequality and therefore also in many papers since the original VAE paper \cite{kingma2013}. Furthermore, beside the assumption that we could in principle observe the latent variables but choose not to, hence the term "unobserved", in general we do not know the exact values. Because of the symmetry of the KL-divergence under diffeomorphisms, this allows for extra functional freedom of the distributions and the involved mapping. Instead of complete freedom, however, it was pointed out in \cite{Khemakhem2020} that the determined final distribution matches the true one up to certain class of transformations $\mathcal{A}$, which they call "identifiability" up to $\mathcal{A}$. In particular, when all terms in the ELBO depend on extra observed input, the prior has a certain structure and the data $x$ are Gaussian observations, $\mathcal{A}$ turns out to be a global scaling and permutation of latent dimensions. \footnote{And potentially a further linear transformation, see \cite{Khemakhem2020} for details.} Regarding the above derivation we can write $\mathcal{R}(\phi^*, \theta^*, \psi^*) \underrel{\mathcal{A}}{\sim}0$ to denote this situation, which means the residual term is zero within the identifiability class $\mathcal{A}$. Extra constraints on the mutual information between data and labels \cite{Zhao2017}, the total correlation of latent variables \cite{Kim2018}, a better prior parametrization \cite{Tomczak2017} or extra conditional dependencies \cite{Khemakhem2020} are often used to improve the ELBO solution. From the above derivation these are all well motivated, as all of those use extra constraints besides the ELBO and therefore have the potential to reduce $\mathcal{R}(\phi^*, \theta^*, \psi^*)$ within a given identifiability class $\mathcal{A}$ whose properties will depend on the constraints. There are also approaches that change the relative strength of the data PDF term  with either a data-PDF or latent-PDF prefactor \cite{Higgins2017} \cite{Zhao2017}. These can be motivated with a balancing of the often very different dimensionality between the two PDFs. For the discussion in the following, we will set this relative scaling to unity without loss of generality.

We can further form the sample approximation of the ELBO and add the above mentioned constraints to form a loss function

\begin{align}
\begin{split}
 &\ \ \ \argmin_{\theta, \phi, \psi}\mathcal{L}_{\mathrm{VAE}}(\theta, \phi, \Psi) \\ 
 &\equiv \argmin_{\theta, \phi, \psi} -\widehat{\mathrm{ELBO}}(\theta,\phi, \psi) + C(\theta, \phi, \psi) 
 \end{split}
 \\
 \begin{split}
&=\argmin_{\theta, \phi, \psi} \frac{1}{N} \sum_{ {x_i,z_{u,i, \phi}}}  - \mathrm{ln}  p_{\theta}(x_i;z_{u,i,\phi}) \\
&\ \ \ \ \ \ \ + \mathrm{ln} \left (\frac{ q_{\phi}(z_{u,i,\phi};x_i)}{ p_\psi(z_{u,i, \phi})}\right)  + C(\theta, \phi, \psi) ,
\end{split}
\end{align}
where the term  $C(\theta, \phi, \psi)$ indicates the constraint. For a total correlation constraint\cite{Kim2018}, for example, we would have  $C(\theta, \phi, \psi) =  \gamma \cdot D_{\mathrm{KL}}(q_{\phi}(z_u); \prod_j q_{j,{\phi}}(z_u))$, where the index $j$ describes the different marginal distributions of $q_{\phi}$. There is also a tunable prefactor $\gamma$ to balance the different loss terms. 
As discussed before, if flexible enough density estimators are used and the final parameter solution is denoted by $\theta^*$, $\phi^*$ and $\psi^*$, it follows that $\mathcal{R}(\phi^*, \theta^*, \psi^*) \underrel{\mathcal{A}}{\sim}0$, the ELBO saturates, and $\mathrm{Surrogate}_{\mathrm{KL}}(\mathcal{P}_t;q_{\phi^*}) \underrel{\mathcal{A}}{\sim} 0$ and $D_\mathrm{KL, joint(x,z_o)} (\mathcal{P}_t;q_{\phi^*}) \underrel{\mathcal{A}}{\sim}0$.

In our view, a few advantages arise from this derivation of the variational autoencoder.
\begin{enumerate}
    \item The derivation is connected to the KL-divergence derivation of supervised learning.
    \item It explicitly shows that there are three joint distributions involved. The first is $\mathcal{P}_t(z_u;x) \cdot \mathcal{P}_t(x)$, the true underlying joint distribution. The second is $q_\phi(z_u;x) \cdot \mathcal{P}_t(x)$, a distribution where the conditional distribution over the latent variables is exchanged for a tractable approximation $q_\phi(z_u;x)$. The third is another tractable approximation $p_\theta(x;z_u) \cdot p_\psi(z_u)$ which is parametrized in the opposite causal direction. In the literature, the true distribution $\mathcal{P}_t$ is typically simply denoted by $p$, going back to the original ELBO marginal likelihood derivation \cite{jordan_elbo_derivation} or VAE publication \cite{kingma2013}. This can be confusing, as $p$ (i.e. $\mathcal{P}_{t}$) and $p_\theta$ are often used interchangeably. 
   
    \item Extra constraints \cite{Zhao2018} \cite{Khemakhem2020} are often invoked in practical VAE training to find a better solution than in vanilla VAE training, but this is not well-motivated in the original VAE derivation. The residual term $\mathcal{R}$ explicitly shows why this practice is useful. It is desired to try to reduce $\mathcal{R}$, and thereby $D_{\mathrm{KL}}(q_{\phi^*}(z_u;x)\cdot\mathcal{P}_{t}(x), \mathcal{P}_{t}(z_u,x))$, within a given identifiability class $\mathcal{A}$ by invoking extra constraints on top of the ELBO loss. 
    
    \item Because of the connection to the supervised loss, there is a natural path to derive the  extended supervised loss as discussed in the next section. This is the most important aspect of this construction for practical applications, because the extended supervised loss allows to calculate a goodness-of-fit (see Section \ref{sec:goodness_of_fit}).
\end{enumerate}

\subsection{Extended supervised and semi-supervised learning}
\label{sec:extended_supervised}

Now let us assume that we have a fixed number of observed and unobserved latent variables $z_o$ and $z_u$, respectively. The KL-divergence of the joint distribution can then be expanded as

\begin{align}
\begin{split}
 &\ \ \ D_{\mathrm{KL, joint(x,z_o,z_u)}} (\mathcal{P}_t;q) \\ 
 &= \int\limits_{x}^{}\int\limits_{z_o}^{}\int\limits_{z_u}^{} \mathcal{P}_t(z_u, z_o,x) \cdot \mathrm{ln} \frac{\mathcal{P}_t(z_u, z_o ,x)}{q(z_u, z_o,x)} dz_u z_o dx 
 \end{split}
 \\ 
 \begin{split}
&=E_{x}\biggl[\int\limits_{z_o}^{} \mathcal{P}_t(z_o;x) \\ 
&\ \ \ \cdot\underbrace{ \textcolor{orange}{\int\limits_{z_u}^{} \mathcal{P}_t(z_u;z_o,x) \mathrm{ln} \frac{\mathcal{P}_t(z_u; z_o ,x)}{q_\phi(z_u; z_o,x)} d z_u} }_{\equiv D_{\mathrm{KL}}(\mathcal{P}_t(z_u;z_o,x);q_\phi(z_u;z_o,x)} d z_o \biggr] \\ &+ \underbrace{E_{x,z_o}\left[ \mathrm{ln} \frac{\mathcal{P}_t(z_o;x)}{q_{\varphi}(z_o;x)} \right] + E_{x}\left[ \mathrm{ln} \frac{\mathcal{P}_t(x)}{q(x)} \right]}_{\equiv \mathcal{S}(\varphi)} \label{eq:surrogate_plus_s},
\end{split}
\end{align}
where the first term involves an intractable KL-divergence similar to the VAE and the second term is the supervised loss function which we abbreviate by $\mathcal{S}$. Next we construct a surrogate term similar to the VAE derivation in order to obtain a tractable objective.

\begin{align}
\begin{split}
&\mathrm{Surrogate}_{\mathrm{KL},1} \\
&=E_{x}\biggl[\int\limits_{z_o}^{} \mathcal{P}_t(z_o;x) \\ 
&\ \ \ \cdot \textcolor{NavyBlue}{\int\limits_{z_u}^{} q_\phi(z_u;z_o,x) \mathrm{ln} \frac{q_\phi(z_u; z_o ,x)}{\mathcal{P}_t(z_u;z_o,x) } d z_u }d z_o \biggr] + \mathcal{S}(\varphi) \label{eq:surrogate_kl1}.
\end{split}
\end{align}
In this term we again replace the KL-divergence (\textcolor{orange}{orange}) by a reverse KL-divergence (\textcolor{NavyBlue}{blue}). In contrast to the VAE, we can define a second surrogate term as 

\begin{align}
\begin{split}
&\mathrm{Surrogate}_{\mathrm{KL},2} \\
&=E_{x}\biggl[\int\limits_{z_o}^{} \textcolor{PineGreen}{\widetilde{q}_\varphi(z_o;x)} \\ 
&\ \ \ \cdot \textcolor{NavyBlue}{\int\limits_{z_u}^{} q_\phi(z_u;z_o,x) \mathrm{ln} \frac{q_\phi(z_u; z_o ,x)}{\mathcal{P}_t(z_u;z_o,x) } d z_u }d z_o \biggr] + \mathcal{S}(\varphi) \label{eq:surrogate_kl2},
\end{split}
\end{align}
where the expectation is taken with respect to the parametrized distribution $\widetilde{q}_\varphi(z_o;x)$ (\textcolor{PineGreen}{turquoise}) that is determined in the supervised part $\mathcal{S}$. The tilde indicates that gradients with respect to $\varphi$ are not propagated through, which is used to completely decouple the supervised from the unsupervised part during learning. This second surrogate loss will be useful for consistent goodness-of-fit procedures (Section \ref{sec:goodness_of_fit}). If flexible enough density estimators are used, we can observe that

\begin{align}
\begin{split}
&\argmin_{\phi, \varphi} {D_\mathrm{KL, joint(x,z_u, z_o)}} (\mathcal{P}_t;q_{\phi/\varphi}) \\
= &\argmin_{\phi, \varphi} \mathrm{Surrogate}_{\mathrm{KL},1}(\phi, \varphi)
\end{split}
\\
= &\argmin_{\phi, \varphi} \mathrm{Surrogate}_{\mathrm{KL},2}(\phi, \varphi)\label{eq:surrogate_extended_equality},
\end{align}
since the supervised part leads to $\widetilde{q}_{\varphi}(z_o;x)\approx \mathcal{P}_t(z_o;x)$
and the surrogate losses have the same global minimum as the joint KL-divergence, similar to the VAE derivation.
\clearpage 
\raggedright{Next, we reformulate the second surrogate loss as}
\begin{strip}
\begin{align}
\begin{split}
 \mathrm{Surrogate}_{\mathrm{KL},2}  &=
 E_{x}\left[ \int\limits_{z_o}^{}  \int\limits_{z_u}^{} \widetilde{q}_\varphi(z_o;x) \cdot q_{\phi}(z_u;z_o,x) \cdot \mathrm{ln} \frac{q_{\phi}(z_u;z_o,x)}{\mathcal{P}_t(z_u;z_o,x)} d z_u d z_o \right] \\
 &\ \ \ + \mathcal{S}(\varphi) \label{eq:semisupervised_surrogate} 
 \end{split}
 \\ 
 \begin{split}
 &= E_{x}\left[ \int\limits_{z_o}^{}  \int\limits_{z_u}^{}  \widetilde{q}_\varphi(z_o;x) \cdot q_{\phi}(z_u;z_o,x) \cdot \mathrm{ln} \frac{q_{\phi}(z_u;z_o,x)\cdot \mathcal{P}_t(z_o,x)}
 {\mathcal{P}_t(z_u,z_o,x)} d z_u d z_o \right] \\
 &\ \ \ + \mathcal{S}(\varphi)
 \end{split}
 \\ \begin{split}&= E_{x}\left[ \int\limits_{z_o}^{}  \int\limits_{z_u}^{}  \widetilde{q}_\varphi(z_o;x) \cdot q_{\phi}(z_u;z_o,x) \mathrm{ln} \frac{q_\phi(z_u;z_o,x) \cdot \mathcal{P}_t(z_o,x)}{p_\theta(x;z_o,z_u)\cdot p_\psi(z_o,z_u)} d z_u d z_o\right] \\ 
 &\ \ \ + \underbrace{E_{x}\left[\int\limits_{z_o}^{}  \int\limits_{z_u}^{}  \widetilde{q}_\varphi(z_o;x) \cdot q_{\phi}(z_u;z_o,x) \mathrm{ln} \frac{p_\theta(x;z_o,z_u)\cdot p_\psi(z_o,z_u)}{\mathcal{P}_t(x,z_o,z_u)}d z_u d z_o \right]}_{\equiv \mathcal{R}(\theta, \phi, \varphi, \psi)} \\
 &\ \ \ + \mathcal{S}(\varphi) 
 \end{split}
 \\
 \begin{split}
 &= E_{x}\left[\int\limits_{z_o}^{} \widetilde{q}_\varphi(z_o;x) \cdot \mathrm{ln}(\mathcal{P}_t(z_o,x)) \  d z_o \right] \\
 &\ \ \ + \underbrace{E_{x}\left[ \int\limits_{z_o}^{}  \int\limits_{z_u}^{}  \widetilde{q}_\varphi(z_o;x) \cdot q_{\phi}(z_u;z_o,x) \mathrm{ln} \frac{q_\phi(z_u;z_o,x)}{p_\theta(x;z_o,z_u)\cdot p_\psi(z_o,z_u)} d z_u d z_o\right]}_{\mathrm{ELBO-like}} \\ 
 &\ \ \ + \mathcal{R}(\theta, \phi, \varphi, \psi) + \mathcal{S}(\varphi)\label{eq:last_semisupervised} ,
\end{split}
\end{align}
\end{strip}
and we end up with a supervised term $\mathcal{S}$ and similar terms to the unsupervised derivation. The residual term $\mathcal{R}(\theta, \phi, \varphi, \psi)$ again is equal to a KL-divergence up to a certain identfiability class after training. There is a slight difference, in that the prior in the ELBO is now defined over the joint space $(z_u, z_o)$ instead of just $z_u$ alone. The same derivation would work with the first surrogate loss by replacing $\widetilde{q}_\varphi(z_o;x)$ with $\mathcal{P}_t(z_o;x)$ everywhere. In particular, if in addition to using the first surrogate loss, the joint prior is split up as $p_\psi(z_o,z_u)=p_\psi(z_u;z_o)\cdot p_\psi(z_o)$, every part of the resulting ELBO objective is conditioned on $z_o$ as an extra parameter. This is an important factor for identifiability guarantees, as outlined in \cite{Khemakhem2020}, and might be interesting to study on its own. We use the second surrogate loss here to later have self-consistent goodness-of-fit calculations (see Section \ref{sec:goodness_of_fit}) at all times during training. 
In the following, we discuss two types of loss functions that can be formed using the previous results.

\paragraph{Extended supervised loss}
The first is a loss definition that can be used to perform what we call \emph{extended supervised training}, and will be important for the calculation of a goodness-of-fit as described in Section \ref{sec:goodness_of_fit}. It is a sample-based application of the tractable parts in eq. \ref{eq:last_semisupervised}: the ELBO-like term and the supervised part $\mathcal{S}(\varphi)$:

\begin{align}
\begin{split}
&\ \ \ \ \argmin_{\theta, \phi, \varphi, \Psi} \mathcal{L}_{\mathrm{ext. supervised}}(\theta, \phi, \varphi, \Psi) \\
&= \argmin_{\theta, \phi, \varphi, \Psi}  \frac{1}{N} \sum_{{x_i,\widetilde{z_o}_{,i,\varphi}, z_{u,i,\phi}}}   \mathrm{ln}(\mathcal{P}_t(\widetilde{z_o}_{,i,\varphi},x_i)) \\ 
&\ \ \ + \mathrm{ln} \frac{q_\phi(z_{u,i,\phi};\widetilde{z_o}_{,i,\varphi},x_i) }{p_{\theta}(x_i;\widetilde{z_o}_{,i,\varphi},z_{u,i,\phi})\cdot p_{\Psi}(\widetilde{z_o}_{,i,\varphi},z_{u,i,\phi})} \\
&\ \ \ + \hat{\mathcal{S}}(\varphi) + C(\theta, \phi, \varphi, \psi) 
\end{split} 
\\
\begin{split}
&=  \argmin_{\theta, \phi, \varphi, \Psi} \frac{1}{N} \\
&\sum_{{x_i,\widetilde{z_o}_{,i,\varphi}, z_{u,i,\phi}}} \underbrace{\mathrm{ln} \frac{q_\phi(z_{u,i,\phi};\widetilde{z_o}_{,i,\varphi},x_i) }{p_{\theta}(x_i;\widetilde{z_o}_{,i,\varphi},z_{u,i,\phi})\cdot p_{\Psi}(\widetilde{z_o}_{,i,\varphi},z_{u,i,\phi})}}_{\mathrm{ELBO-like}} \\
&\ \ \ + \underbrace{\hat{\mathcal{S}}(\varphi)}_{\mathcal{L}_{\mathrm{supervised}}} + C(\theta, \phi, \varphi, \psi).
\label{eq:extsupervised_loss} 
\end{split}
\end{align}
We again add a constraint term $C(\theta, \phi, \varphi, \psi)$ similar to the VAE loss to potentially improve the identifiability of the unsupervised dimensions $z_u$. 
The true observed labels are denoted by $z_{o,i}$, samples from $\widetilde{q}_\varphi(z_o;x)$ are denoted by $\widetilde{z_o}_{,i,\varphi}$. The symbol $\sim$ indicates that the gradient is not propagated in order to decouple the supervised part during training. If the other surrogate term $\mathrm{Surrogate}_{\mathrm{KL},1}$  (eq. \ref{eq:surrogate_kl1}) had been used, this decoupling would have happened automatically. Because the supervised training is effectively decoupled, one can also choose to first train the supervised part $\mathcal{L}_{\mathrm{supervised}}$, and only later train the rest. An extended supervised training can therefore always be started with an already finished supervised model and can be viewed as an add-on to it.

\paragraph{Semi-supervised learning}

The extended supervised loss can also be adapted for semi-supervised learning. In semi-supervised learning, parts of the training data have labels, and parts are unlabeled. In the derivation of the VAE loss we argued that latent variables that share mutual information with the data can in principle be labeled. In semi-supervised learning, this assumption is automatically implied - parts of the data are not labeled, but could be in principle, if the data comes from a Monte Carlo simulation. Taking the variational viewpoint of the previous sections, a naive solution might be to use the supervised loss for data with labels, the VAE-loss for data without labels, and use the same distribution $q_\phi$ in both losses to share parameters. A more natural solution, however, is to use the extended supervised loss instead of the pure supervised loss. 

\begin{align}
 &\argmin_{\theta, \phi, \varphi, \Psi} \mathcal{L}_{\mathrm{semi-supervised}}(\theta, \phi, \varphi, \Psi) \\ =  &\argmin_{\theta, \phi, \varphi, \Psi} \left\{
  \begin{array}{@{}ll@{}}
    \displaystyle \mathcal{L}_{\mathrm{ext. supervised}}(\theta, \phi, \varphi, \Psi) & \ (\mathrm{labeled} \ \mathrm{data}) \\
    \mathcal{L}_{\mathrm{VAE}}(\theta, \phi, \varphi, \Psi) & \ (\mathrm{unlabeled} \ \mathrm{data})
  \end{array}\right.
\end{align}

For unlabeled data, we take the loss of the variational autoencoder over the combined space $z_{\mathrm{comb}}=\lbrace z_o,z_u \rbrace$ and treat the combined variable as unsupervised.
The parametrization does not change for data with or without labels, only the sampling strategy differs. Therefore, the Ansatz seems to be an elegant and natural way to unify both types of data.

\section{Toy Monte Carlo}
\label{sec:toy_mc}

\begin{figure*}
    \centering
    \begin{subfigure}[t]{0.48\textwidth}
    	\includegraphics[width=\textwidth]{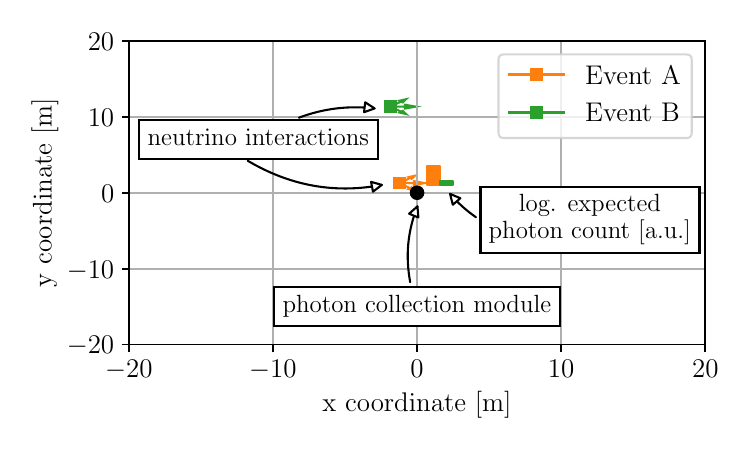}
    	\caption{} 
    \end{subfigure}
    \begin{subfigure}[t]{0.48\textwidth}
    	\includegraphics[width=\textwidth]{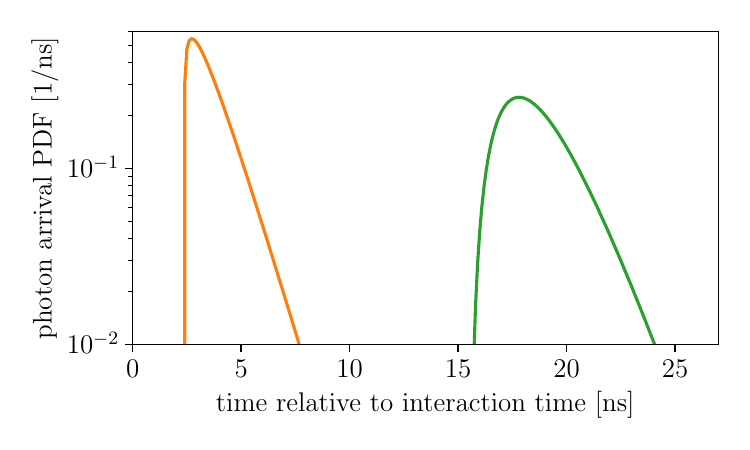}
    	\caption{} 
    \end{subfigure}
    \begin{subfigure}[t]{0.48\textwidth}
    	\includegraphics[width=\textwidth]{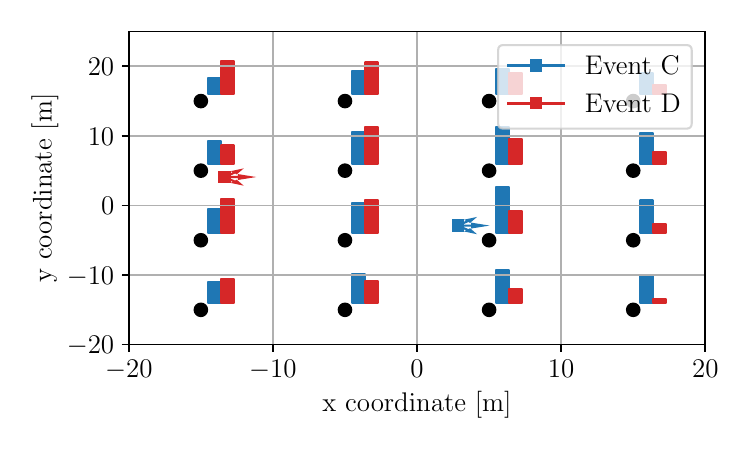}
    	\caption{} 
    \end{subfigure}
    
    \caption{Illustration of the two simplest toy Monte Carlo datasets. Black dots denote collection modules, squares with arrows indicate shower-like neutrino events and vertical bars expected logarithmic photon yield in a given photodetector. (a) Dataset 1 (single photodetector) with two example events A and B. (b) Photon arrival time distributions of events A and B. (c) Dataset 2 (16 photon collectors) with two example events C and D. }
\label{fig:toy_mc}
\end{figure*}

Several toy Monte Carlo datasets have been created to perform empirical tests in the following sections. They are designed to mimic electromagnetic showers of electron-neutrino interactions \cite{pdg_2019} in a Cherenkov neutrino detector like IceCube \cite{Aartsen2017} in a 2-D setting. In reality, such showers consist of charged particles that emit Cherenkov light in a coherent light front at the Cherenkov angle that changes its shape as detection modules are further away from the interaction point \cite{radel_2012}. For ice, which is the detection medium in IceCube, the light front becomes nearly isotropic for large distances. In general, depending on the position and orientation of the shower with respect to a detection module, the shape of the resulting arrival time distribution of photons varies. In this toy simulation the arrival time distribution is parametrized by a gamma distribution. In addition, the expected number of detected photons falls off exponentially with distance and depends on the orientation of the shower with respect to the module. These PDFs are parametrized such that they qualitatively mimic the photon PDF behavior of the real IceCube detector\cite{Aartsen2017}. At detection, the actual number of observed photons follows a Poisson distribution. Fig. \ref{fig:toy_mc} illustrates the first two datasets that are used in the next sections. 

Fig. \ref{fig:toy_mc} (a) shows a detector that consists of a single module with two example neutrinos \emph{A} and \emph{B}. The two bars next to the photon collection module indicate the logarithmic expected mean of the number of observed photons from each neutrino interaction. Fig. \ref{fig:toy_mc} (b) shows the two corresponding photon arrival distributions. For event \emph{A}, the data distribution is more concentrated than for event \emph{B}, because the particle interacts closer to the detection unit. Also the number of observed photons is larger. Fig. \ref{fig:toy_mc} (c) shows the configuration of a second set of Monte Carlo simulations for a larger detector and a simulated threshold condition that at least 5 photons are observed for an event to be recorded. Also depicted are two associated example events (\emph{C} and \emph{D}) together with their expected number of photons in the various photon collection modules. 

A summary of all datasets is given in table \ref{table:datasets}.
Dataset 1 and 2 have two intrinsic degrees of freedom, the position of each neutrino interaction. The deposited energy and thereby emitted photons is always the same. Dataset three has an additional degree of freedom by also randomizing the direction. Dataset four also simulates a falling energy spectrum between $1$ and $100$ GeV. Datasets five to seven involve a larger detector and also four degrees of freedom. The last dataset, which contains track-like topologies, effectively emulates relativistic muon tracks by putting multiple eletromagnetic showers along a track that moves with the speed of light and distributing the energy in equal parts among those losses.  More information on how these more complex datasets are used is given in Section \ref{sec:goodness_of_fit}.

The detection process can be described by an inhomogeneous tempo-spatial Poisson point process \cite{poisson_process_book}. The spatial part is restricted to the position of the collection modules, while the temporal detection can happen at all times. The corresponding likelihood function for observed labels $z_o$ is an extended likelihood \cite{Barlow1990}, $\mathcal{L}(z_o)$, which we can write as an explicit evaluation of the data generating distribution $\mathcal{D}(\textbf{k},\textbf{t};z_o,x)$. For a total number of modules $N$, this can be written as
\begin{align}
\begin{split}
\mathcal{L}(z_o) &= \mathcal{D}(\textbf{k},\textbf{t};z_o,\textbf{x}) \\ 
&=\prod\limits_{j=1}^{N} \frac{\mathrm{exp}^{-{\lambda_j(z_o,\textbf{x}_j)} }\cdot {\lambda_j(z_o,\textbf{x}_j)}^{k_j} }{k_j!} \prod\limits_{i=1}^{k_j} p_j(t_i;z_o,\textbf{x}_j) \label{eq:likelihood},
\end{split}
\end{align} 
where $\lambda_j$ is the expectation value of a Poisson distribution for module $j$, $k_j$ the detected number of photons in module $j$, and $p_j(t_i)$ the probability distribution of photon arrival time $t_i$ in module $j$. The Poisson mean $\lambda_j$ and the shape of $p_j(t_i)$ depend on the event parameters $z_o$ and module positions $\textbf{x}_j$.

\begin{table*}

\begin{tabular}{ |c|c|c|c|c| } 
 \hline
  & no. modules & d.o.f. & event & other \\  \hline 
 dataset 1 & 1 &  2 (pos.) & shower-like & $> 1$ obs. photon  \\ 
 dataset 2 & 16 &  2 (pos.) & shower-like & $> 5$ obs. photons  \\ 
 dataset 3 & 16 &  3 (pos.+dir.) & shower-like & $> 5$ obs. photons  \\ 
 dataset 4 & 16 &  4 (pos.+dir.+energy) & shower-like & $> 5$ obs. photons, $0-50 \%$ light-yield   \\ 
 dataset 5 & 400 &  4 (pos.+dir.+energy) & shower-like & $> 5$ obs. photons, $|x|,|y|<55$m, $E\propto E^{-1}$  \\ 
 dataset 6 & 400 &  4 (pos.+dir.+energy) & shower-like & $> 5$ obs. photons, $E\propto E^{-1}$  \\ 
 dataset 7 & 400 &  4 (pos.+dir.+energy) & track-like & $> 5$ obs. photons, $E\propto E^{-1}$  \\ 
 \hline
\end{tabular}

\caption{Properties of the datasets used for various comparisons. The detector geometry always consists of 1 or 16 modules as depicted in Fig. \ref{fig:toy_mc} or of 400 modules as depicted in Fig. \ref{fig:gof_supplemantary}}
\label{table:datasets}
\end{table*}

\section{The importance of flow-based models}
\label{sec:importance_flow_based_models}

All derivations so far assumed non-specific parametrizations of conditional probability density functions, i.e. posteriors and data generating PDFs, with neural networks. A general way to parametrize a probability density function with a neural network is via conditional amortized normalizing flows \cite{amortized_nfs} \cite{nf_review}. Normalizing flows are defined using a flow-defining function $\rho_{\vec{F}}(\hat{z})$ whose parameters we define as $\vec{F}$. This function can be used as $z_o=\rho_{\vec{F}}(\hat{z})$ to transform a base random variable $\hat{z}$, usually following a standard normal distribution, to the desired target random variable $z_o$. For normalizing flows to work, this function has to be invertible and differentiable, i.e. it has to be a diffeomorphic. If these properties are satisfied, then denoting the probability density  of the base by $p_b(\hat{z})$ and the target by $q(z_o)$ we can calculate the log-probability of the target 

\begin{align}
\mathrm{ln}\big(q(z_o)\big)=\mathrm{ln}\big(p_b(\rho_{\vec{F}}^{-1}(z_o))\big)-\mathrm{ln} \big(\mathrm {det} (J^{\rho_{\vec{F}}}_{\hat{z}})\big) \label{eq:nf_definition},
\end{align}
where $J^{\rho_{\vec{F}}}_{\hat{z}}$ is the Jacobian of $\rho_{\vec{F}}$ with respect to $\hat{z}$. Flows can be composed of multiple other flows $\rho=\rho_1 \circ \rho_2 \dots \circ \rho_n$ and the resulting log-probability simply involves a sum over all log-determinants. In general $p_b(\hat{z})$ can be arbitrary, but for simplicity and the possibility to calculate coverage (see Section \ref{sec:coverage}) we use the standard normal distribution. One can also generate samples from $q(z_o)$ by first sampling $\hat{z}_i$ from $p_b(\hat{z})$ and then transform the samples via $z_{o,i,\vec{F}}=\rho_{\vec{F}}(\hat{z}_i)$, where the samples now depend on the flow parameters $\vec{F}$. This makes the samples differentiable, known as the reparametrization trick \cite{kingma2013}, and is a key feature in the variational autoencoder and extended supervised losses (Section \ref{sec:extended_supervised}). 

\paragraph{Conditional normalizing flows}
Standard normalizing flows only describe PDFs without conditional dependencies, but we would like them to describe conditional PDFs like the posterior distribution. There are in principle multiple ways how this can be achieved. One way is to extend specific normalizing flows designed for high-dimensional image data, like NICE \cite{nice} and GLOW \cite{glow}. These type of flows contain neural-network conditioners, typically MLPs, as part of their flow definition, and one can add a data representation as an additional input to these conditioners. This has been done before, e.g. in \cite{cinns_2021}. There are another class of normalizing flows which are parameter-efficient Euclidean normalizing flows without coupling layers like "radial flows" \cite{amortized_nfs} or "Gaussianization flows" \cite{Meng2020} that have been shown in low Euclidean dimension ($D \lessapprox 20$) to have a good performance on density estimation. This is the dimensionality regime we are interested in. These types of flows also allow a natural way to create a conditional normalizing flow by just predicting all the flow parameters by a neural network, which is the strategy we follow in this paper.

The way to describe such a conditional normalizing flow $q(z_o;x)$ with conditional input $x$ is to make the transformation $\rho_{\vec{F}}$ dependent on $x$ via a non-linear neural network transformation, as indicated in Fig. \ref{fig:nf_illustration} (a). In general, a neural network with parameters $\phi$ that takes $x$ as an input predicts the parameters $\vec{F}$, which in turn defines $\rho_\phi(\hat{z};x)\equiv\rho_{\vec{F}_\phi(x)}(\hat{z})$. The log-probability of a conditional normalizing flow PDF then looks like

\begin{align}
\mathrm{ln}(q_\phi(z_o;x))=\mathrm{ln}(p_b(\rho^{-1}_\phi(z_o;x)))-\mathrm{ln} (\mathrm {det} (J^{\rho_\phi(\hat{z};x)})).\label{eq:nf_definition_conditional}
\end{align}
Compared to a standard normalizing flow (eq. \ref{eq:nf_definition}) which has flow parameters $\vec{F}$, the free parameters of such a conditional normalizing flow are actually the neural network parameters $\phi$ of the network used to encode the data $x$.
As a specific example, Fig. \ref{fig:nf_illustration} (b) shows an affine flow where a neural network predicts a mean vector $\bar{\mu}$ and a width $\sigma$ when the base distribution is a standard normal distribution. The resulting probability distribution $q_\phi(z_o;x)$ of this flow describes a symmetric Gaussian distribution with mean $\bar{\mu}$ and standard deviation $\sigma$. This is a common choice in some regression problems in high-energy neutrino physics \cite{km3net_ev_reco} and used later in some comparisons. 

The most common choice in supervised learning is to fix $\sigma=1$, which results in a Gaussian PDF with unit variance which corresponds to the Mean-Squared-Error (MSE) loss function. Generic conditional normalizing flows with a standard normal distribution as base distribution therefore naturally generalize the MSE-loss. 

\paragraph{Normalizing flows on spheres and tensor products of manifolds}

Normalizing flows can be defined on manifolds like 1-spheres ($\mathbb{S}^1$) or 2-spheres ($\mathbb{S}^2$). In physics, 2-spheres are in particular interesting because directions in space are naturally defined on the 2-sphere, in particular if one wants to avoid issues in the polar regions which come up when looking at the zenith and azimuth seperately. Manifolds of dimension $n$ are always embedded in an Euclidean embedding space $\mathbb{R}^{n+1}$. In general, as described in \cite{Rezende2020}, the log-determinant factors from eq. \ref{eq:nf_definition} and eq. \ref{eq:nf_definition_conditional} of manifold flows differs from Euclidean flows as

\begin{align}
    \begin{split}
    &\mathrm{ln}\big(\mathrm{det}(J_{\mathrm{Eucl.}})\big) \\
    &\rightarrow \mathrm{ln}\left(\sqrt{\mathrm{det}(E^T\cdot J_{\mathrm{Emb.}}^T \cdot J_{\mathrm{Emb.}} \cdot E}\right),
    \label{eq:manifold_density_update}
    \end{split}
\end{align}
where the Jacobian $J_\mathrm{Emb.}$ is calculated as if the transformation acts in embedding coordinates and an additional orthonormal projection matrix $E$ projects into the tangent space at the transformation coordinates. If the Jacobian corresponds to an Euclidean transformation, the formula reduces to the standard case. In the end, manifold normalizing flows work similar to Euclidean normalizing flows, in that one can calculate the log-probability and sample efficiently from the distribution. The base distribution $p_0$ also has to be defined on the manifold, and for spheres a typical choice is the uniform distribution \cite{Rezende2020}. In section \ref{sec:coverage}, we argue that it is actually useful to have a fixed transformation from the Gaussian distribution in the Euclidean plane to the uniform distribution on the sphere as an additional step to facilitate coverage calculation.

In order to describe distributions that are defined simultaneously over the direction and position of an interaction, we can define a normalizing flow over a product space of manifolds. In the example of a 2-dimensional position and 1-dimensional direction, which appears in the toy simulation in Section \ref{sec:toy_mc}, this space would be $\mathcal{R}^2 \times \mathbb{S}^1$, i.e. 2-dimensional Euclidean space and a 1-sphere. We use an autoregressive structure similar to \cite{iaf} to connect the PDFs over the different manifolds and create a joint PDF on the tensor product space. In order to still use a single combined Gaussian base distribution, we again employ some transformations from the plane to the sphere for all manifold sub-parts as described in Section \ref{sec:coverage}. This allows to calculate coverage also for such tensor product distributions. 

\paragraph{Architecture and training}

In all further comparisons in this paper, we split the parameters of flow-based posteriors into two parts. The first part consists of an encoder with gated recurrent units (GRUs) \cite{Cho2014} and a subsequent aggregate MLP to encode the data into an internal representation $\bm{h}$. The GRU reads in a time-ordered sequence of photons, where each photon is characterized by its detected position and time. The second part is a multi-layer perceptron with 2 layers that further maps that internal representation to the respective flow parameters $\bar{F}$. The process is illustrated in Fig. \ref{fig:nf_illustration} (c). The generative model used in section \ref{sec:goodness_of_fit} uses label and latent variables as an input to an MLP that maps to the flow parameters.

For the training of conditional normalizing flows in supervised learning, for example in section \ref{sec:example_gaussianization_flows}, we evaluate the log-probability of the labels for a given input \ref{eq:nf_definition_conditional} and minimize the negative log probability as defined in eq. \ref{eq:real_supervised_loss} in batches using stochastic gradient descent.

For extended supervised training, which is used for the goodness-of-fit calculation in section \ref{sec:goodness_of_fit}, we additionally learn a generative model for the data and an additional posterior over a latent space as defined in the loss function \ref{eq:extsupervised_loss}. The additional latent space posterior and generative model are trained using the extra ELBO term. We train the supervised loss and the ELBO at the same time in batches using stochastic gradient descent, but make sure that gradients are strictly separated, which corresponds to the "second surrogate loss" described in section \ref{sec:extended_supervised}.

At the end of training we adopt stochastic weight averaging (SWA) \cite{swa} in all cases. We found this to reduce the fluctuations and find a more stable solution. More details on the architecture and on the training procedure is given in appendix \ref{appendix:impl_details}. 

\paragraph{Computational efficiency}

One of the advantages of conditional normalizing flows compared to classical likelihood approaches is its computational efficiency. Astronomical alerts sent by the IceCube detector, for example, often take hours from the time the neutrino enters the detector until the alert is sent to the community as an astronomical telegram (ATEL). The reason are time-intensive profile likelihood scans that are needed for uncertainty contours. Spherical conditional normalizing flows, on the other hand, can produce a full-sky scan in seconds using multi-resolution HEALPIX \cite{healpix} grids because they can both sample and evaluate the PDF. In a first step, samples drawn from the normalizing flow define the grid by guiding where pixelation has to be finer, and where it can be coarser. In a second step the PDF is then evaluated with the multi-resolution grid found in step one. With the other properties described in this paper, like systematics inclusion and coverage guarantees, this makes it an appealing alternative to likelihood scans that take hours. The only time consuming aspect is the training, which can take a few weeks on a single GPU for more complex models. This is of course not really an issue, as the model only needs to be trained once and can then readily be used for inference.

\begin{figure*}
    \centering
    \begin{subfigure}[t]{0.5\textwidth}
        \centering
        \begin{tikzpicture}[
title/.style={font=\fontsize{6}{6}\color{black!50}\ttfamily},
typetag/.style={rectangle, draw=black!50, font=\scriptsize\ttfamily, anchor=west},
node distance=1.3cm,>=latex']
	\node (normal) [circle, draw] {};
	\node (normal_eq) [right of=normal] {$\mathcal{N}(\hat{z};\textbf{0},\mathbbm{1})$};
	
	
	\node(outer_normal) [fit=(normal)(normal_eq), draw, rounded corners] {};

	\node (posterior_eq) [below of=normal_eq, yshift=-1cm] {$q_{\phi}(z_o;x)$};
	\draw [] plot [smooth cycle] coordinates { ($(posterior_eq.west)+(-0.5,0.2)$) ($(posterior_eq.west)+(-0.55,0.15)$) ($(posterior_eq.west)+(-0.45,-0.1)$) ($(posterior_eq.west)+(-0.5,-0.3)$) ($(posterior_eq.west)+(-0.2,-0.2)$) ($(posterior_eq.west)+(-0.05,-0.2)$) ($(posterior_eq.west)+(-0.1,0.1)$) };
	\node (posterior_shadow) [left of=posterior_eq,text width=0.5cm,circle] {};
	\node (outer_posterior) [fit=(posterior_shadow)(posterior_eq), draw, rounded corners] {};
	
	\path[->] (normal) edge[bend left] node (phi_node) [left] {${\rho}_\phi(\hat{z};x)$} (posterior_shadow);

	\node (x) [circle, left of=phi_node, xshift=-1.5cm] {$x$};
	\draw[->] (x) edge[bend right] node (F) [below] {$\vec{F}_\phi(x)$} (phi_node);

\end{tikzpicture}
        \caption{General conditional normalizing flow}
    \end{subfigure}%
    ~ 
    \begin{subfigure}[t]{0.5\textwidth}
        \centering
        \begin{tikzpicture}[
title/.style={font=\fontsize{6}{6}\color{black!50}\ttfamily},
typetag/.style={rectangle, draw=black!50, font=\scriptsize\ttfamily, anchor=west},
node distance=1.3cm,>=latex']
	\node (normal) [circle, draw] {};
	\node (normal_eq) [right of=normal] {$\mathcal{N}(\hat{z};\textbf{0},\mathbbm{1})$};
	
	\node(outer_normal) [fit=(normal)(normal_eq), draw, rounded corners] {};

	\node (posterior_eq) [below of=normal_eq, yshift=-1cm, text width=2.8cm] {$\mathcal{N}(z_o;\overline{\mu}_{\phi},\sigma_{\phi}^2\cdot\mathbbm{1})$};
	\node (posterior_shadow) [left of=posterior_eq, xshift=-0.8cm, text width=0.5cm,circle,draw] {};
	\node (outer_posterior) [fit=(posterior_shadow)(posterior_eq), draw, rounded corners] {};
	
	\path[->] (normal) edge[bend left] node (phi_node) [left] {$\sigma_{\phi}(x) \cdot \hat{z} + \overline{\mu}_{\phi}(x)$} (posterior_shadow);

	\node (x) [circle, left of=phi_node, xshift=-1.5cm] {$x$};
	\draw[->] (x) edge[bend right] node (F) [below] {$\begin{pmatrix}
  \overline{\mu}_{\phi}(x) \\
  \sigma_{\phi}(x) \\
\end{pmatrix}$} (phi_node);
   
\end{tikzpicture}
        \caption{Affine conditional flow with single scaling}
    \end{subfigure}
    
    \begin{subfigure}[t]{0.5\textwidth}
        \centering
        \begin{tikzpicture}[
title/.style={font=\fontsize{6}{6}\color{black!50}\ttfamily},
typetag/.style={rectangle, draw=black!50, font=\scriptsize\ttfamily, anchor=west},
node distance=1.3cm,>=latex']
	
	\node (x) [circle] {$x=\begin{pmatrix}
  x_1,y_1,t_1 \\
  \ldots \\
  x_n,y_n,t_n
\end{pmatrix}$};
	\node (x_hid) [circle, right of=x] {};
	\node (h) [circle, right of=x_hid] {$\bm{h}$};
	\node (F) [circle, right of=h] {$\bm{\vec{F}}$};
	
	\draw[->] (x_hid) edge[bend right, align=center] node (r) [below] {GRU \\ + \\ aggreg. MLP} (h);
	\draw[->] (h) edge[bend right] node (a) [below] {MLP} (F);
    
\end{tikzpicture}
        \caption{Common data encoding for all experiments.}
    \end{subfigure}
    \caption{General conditional flow (a) and an affine conditional flow (b) parametrization of the approximate posterior in supervised learning. Choosing $\sigma_{\phi}=1$ yields a shifted standard normal distribution which is the PDF used in the MSE loss. General normalizing flow parameters are denoted by $\vec{F}$ and the parameters of the encoding neural network are denoted by $\phi$. The common encoding scheme for all experiments is depicted in (c), which amortizes the parameters $\vec{F}$.}
   \label{fig:nf_illustration}
\end{figure*}

\begin{figure*}
    \centering
    \begin{subfigure}[t]{0.48\textwidth}
    	\includegraphics[width=\textwidth]{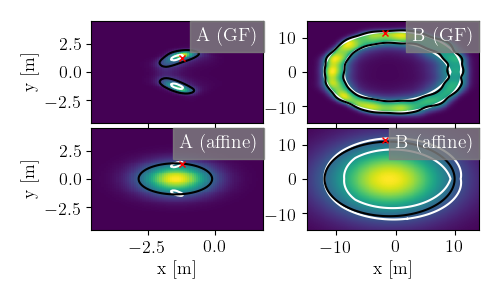}
    	\caption{Posterior scans of dataset 1 example events.} 
    \end{subfigure}
    \begin{subfigure}[t]{0.48\textwidth}
    	\includegraphics[width=\textwidth]{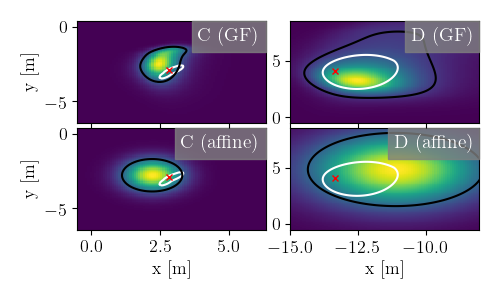}
    	\caption{Posterior scans of dataset 2 example events.} 
    \end{subfigure}
    \caption{A comparison of posteriors of the position for the example events A and B from dataset 1 and events C and D from dataset 2. The normalizing flow posterior is shown together with a $68 \%$ contained probability mass in black. The $68 \%$ probability mass contour of the true posterior assuming a flat prior is shown in white. The true event positions are marked in red. The upper row shows the result for Gaussianization flows, the lower row for an affine flow (a Gaussian) with a single covariance parameter.}
\label{fig:posterior_comparisons}
\end{figure*}

\begin{figure*}
    \centering
    \begin{subfigure}[t]{0.48\textwidth}
    	\includegraphics[width=\textwidth]{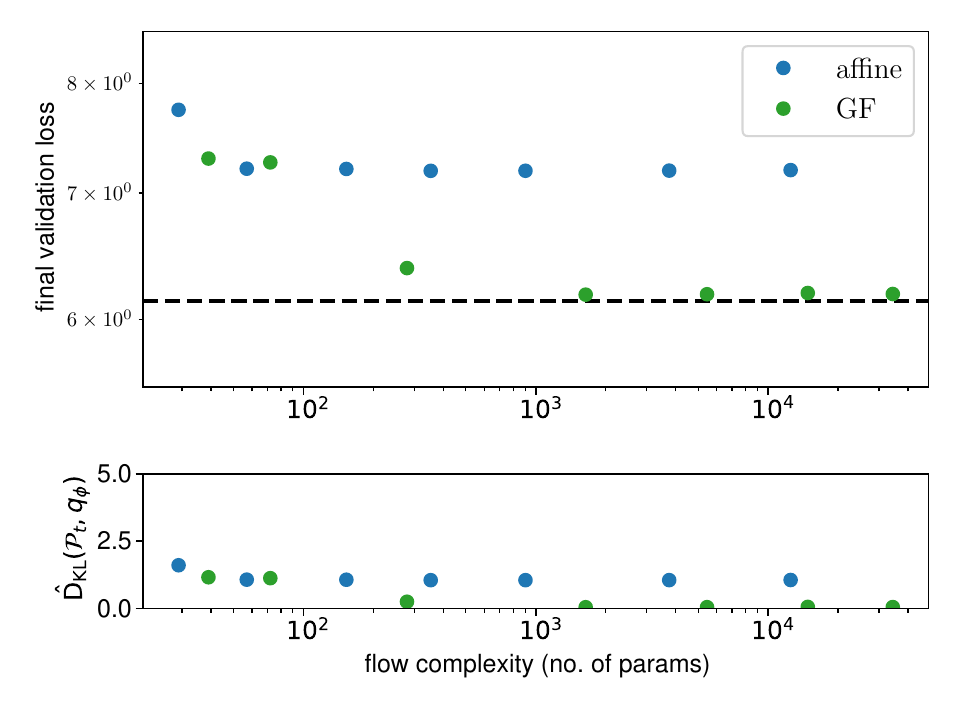}
    	\caption{Performance vs. complexity for dataset 1.} 
    \end{subfigure}
    \begin{subfigure}[t]{0.48\textwidth}
    	\includegraphics[width=\textwidth]{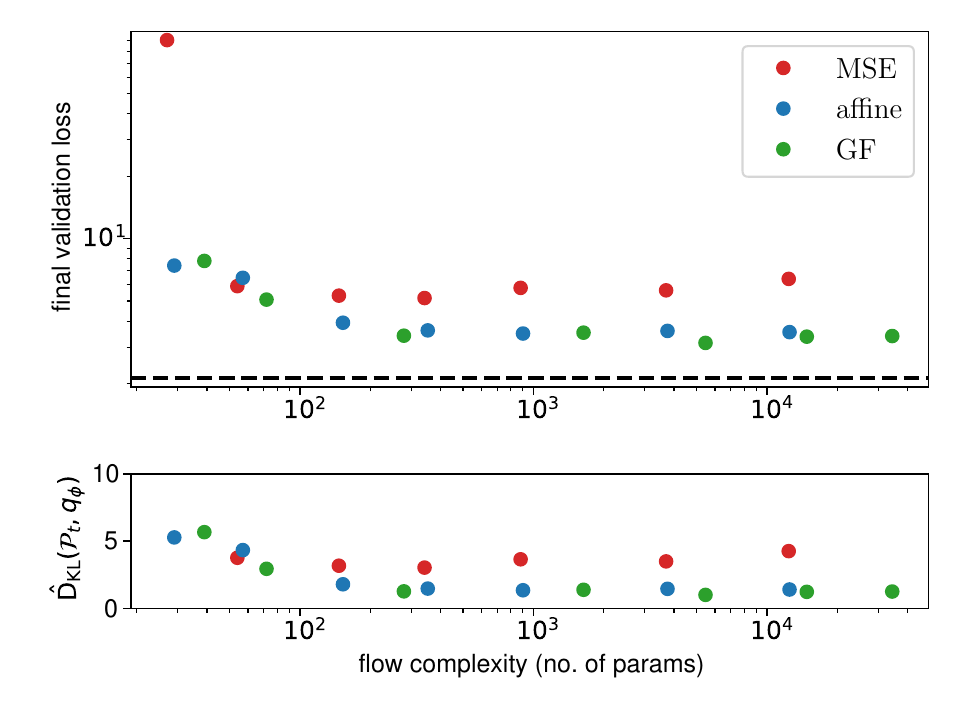}
    	\caption{Performance vs. complexity for dataset 2.} 
    \end{subfigure}
    \caption{Posterior approximation performance of a Gaussianization flow \cite{Meng2020} (GF), an affine flow with a single variable width $\sigma$ (affine), and an affine flow with $\sigma=1$ (MSE). Along the x-axis the number of parameters and the potential flow complexity increases, while the encoding complexity is held fixed (see text). The respective upper plot shows the validation loss. The black dotted line shows the loss obtained from the true posterior. The respective lower plot shows the sample-based KL-divergence. }
\label{fig:posterior_vs_complexity}
\end{figure*}

\section{An example application with conditional Euclidean flows}
\label{sec:example_gaussianization_flows}

In the following we study three Euclidean conditional normalizing flows in a simple example to infer the position of a neutrino interaction in the toy Monte Carlo. With the true likelihood of the true data generating function (eq. \ref{eq:likelihood}) and the inherent prior distribution we can construct a true posterior distribution. We assume a flat prior for simplicity. We can then compare the true posterior with the posterior approximations obtained using various normalizing flows after training. Fig. \ref{fig:posterior_comparisons} shows such a comparison for the example events from section \ref{sec:toy_mc}. It compares a flexible Gaussianization flow \cite{Meng2020} with an affine flow (see Fig. \ref{fig:nf_illustration} for details). For dataset 1, which consists of a single photon collection module, the resulting posterior is highly non-Gaussian. The flexible Gaussianization flow can approximate the posterior much better than an affine flow with a single width parameter. For dataset 2 the events generally contain more observed photons and the posteriors become more Gaussian-like as dictated by the Bernstein-von-Mises theorem \cite{vaart_1998}. Both flows have more comparable shapes in this example. To assess performance more generally, Fig. \ref{fig:posterior_vs_complexity} shows the posterior approximation quality versus the number of parameters of the second part of the flow. The number of parameters of the first GRU encoder part is held fixed and not included in the quantity displayed on the x-axis. The second part, which consists of an MLP with two layers, is varied in its hidden dimension. For the Gaussianization flow, additionally the number of flow parameters is varied, which specifically for Gaussianization flows leads to more internal layers and more kernel basis elements (see appendix \ref{appendix:impl_details} for details). In general, the flexible Gaussianization flow is able to reach better density estimation by having a lower KL divergence to the true distribution than the simpler affine flow. As can be seen in Fig. \ref{fig:posterior_vs_complexity} b), the MSE loss with a standard normal posterior is roughly the same scale as the other posterior approximations in dataset 2, and the corresponding KL-divergence to the true posterior only a little worse than for an affine or Gaussianization flow. This happens by chance here, since true posteriors have various shapes and scales and no reason to match a standard normal distribution. For dataset 1 (Fig. \ref{fig:posterior_vs_complexity} a)), on the other hand, the standard normal using the MSE-loss is much worse in terms of KL-divergence to the true posterior and therefore not shown in the figure. The better approximation performance of the Gaussianization flow has to do with the possibility to adjust the potential complexity of the flow itself, i.e. increase the size of $\bar{F}$. For an affine flow, on the other hand, $\bar{F}$ is fixed to be a mean $\bar{\mu}$ and width $\sigma$, and one can only increase the complexity by increasing the MLP hidden dimension. In all cases, the final performance saturates, where more parameters do not help performance. 

The results suggest that flow approximation capabilities are not the bottleneck in this toy study. As mentioned earlier and indicated in Fig. \ref{fig:posterior_comparisons} a), for a single detection module the posterior shapes are highly non-Gaussian. Yet the Gaussianization flow has a good approximation quality, i.e. a KL-divergence close to zero (Fig. \ref{fig:posterior_vs_complexity} a). In dataset 2, which consists of four detection modules, the posterior shapes tend to be more Gaussian due to the larger amount of photons per event (Fig. \ref{fig:posterior_comparisons} b), so a close posterior approximation should be easier to achieve than for dataset 1 when using the same normalizing flow. The opposite is the case, however, as seen in the slightly higher KL-divergence offset from zero (Fig. \ref{fig:posterior_vs_complexity} b). These results suggest that either more training data is required for this more complex dataset or the encoding architecture needs to be improved - or a combination of both. We come back to this at the very end of the paper.

\section{Coverage of the neural network posterior}
\label{sec:coverage}

In Frequentist or Bayesian analyses it is important to know how well confidence or credible intervals cover the true values. In a Frequentist analysis, in the limit of large amounts of data,  Wilks' Theorem \cite{Wilks1938} implies that the quantity $\lambda=-2\cdot(\mathrm{ln}L(\theta_t)-\mathrm{ln}L(\theta_0))$ is $\chi^2_\nu$ distributed with $\nu$ corresponding to the dimensionality of $\theta$ when $\theta_0$ is the value that maximizes the likelihood function and $\theta_t$ the true value that generated the data. It is connected to the fact that likelihood scans around the optimum have an approximately Gaussian shape in the large-data limit. A similar statement appears in a Bayesian analysis from the Bernstein-von-Mises theorem \cite{vaart_1998}, which implies that for large amounts of data the posterior $p(\theta;x)$ becomes Gaussian. For any n-dimensional multivariate Gaussian the quantity $(x-\mu)C^{-1}(x-\mu)$ is $\chi_n^2$ distributed \cite{chi2_fact}, which again implies  the quantity $\lambda_{\mathrm{Bayes}}=-2\cdot(\mathrm{ln}p(\theta_t;x)-\mathrm{ln}p(\theta_0;x))$ is also $\chi^2$ distributed in the Gaussian limit of $p(\theta;x)$. 
\begin{figure*}
    \centering
    \begin{subfigure}[t]{0.98\textwidth}
    \includegraphics[width=\textwidth]{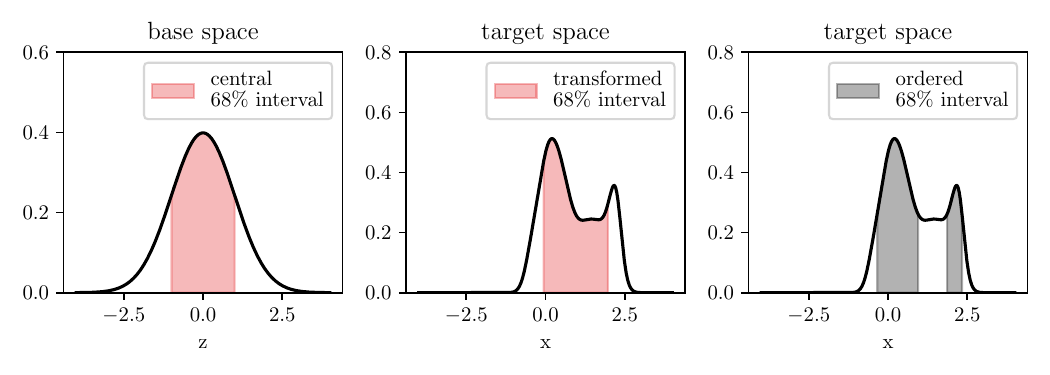}
    \caption{1-d interval illustration}
    \end{subfigure}
    \begin{subfigure}[t]{0.98\textwidth}
    \includegraphics[width=\textwidth]{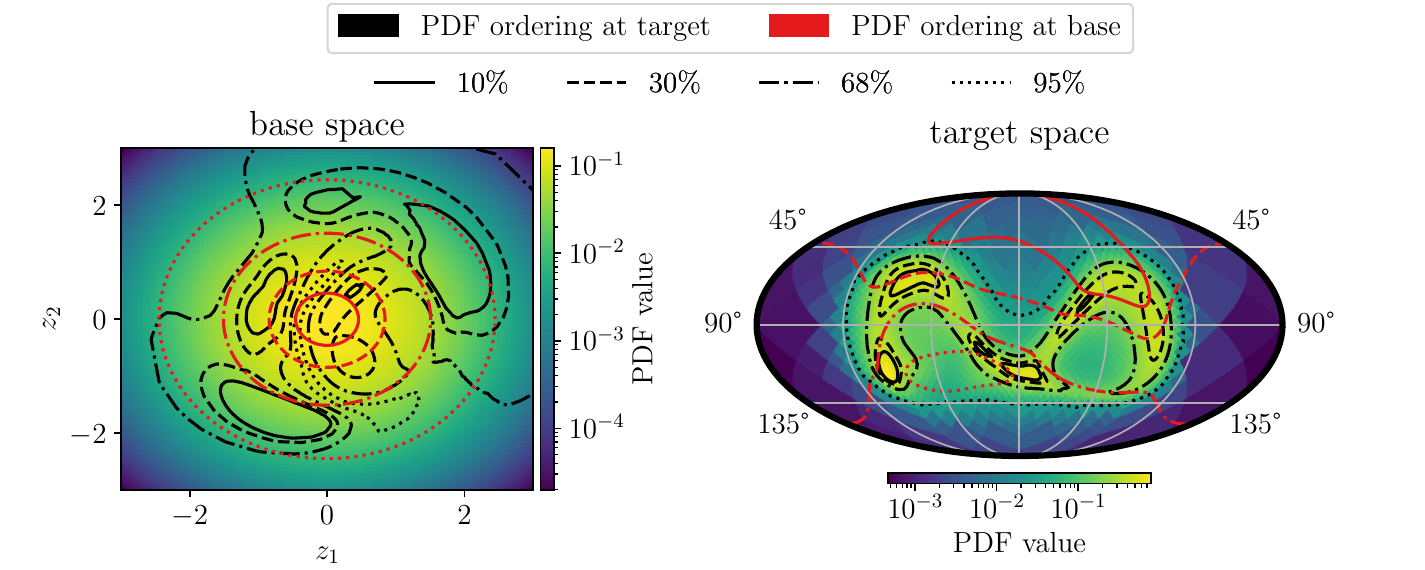}
    \caption{"M" shape on sphere - base-ordered contour is not aligned with target-ordered contour}
    \end{subfigure}
    \begin{subfigure}[t]{0.98\textwidth}
    \includegraphics[width=\textwidth]{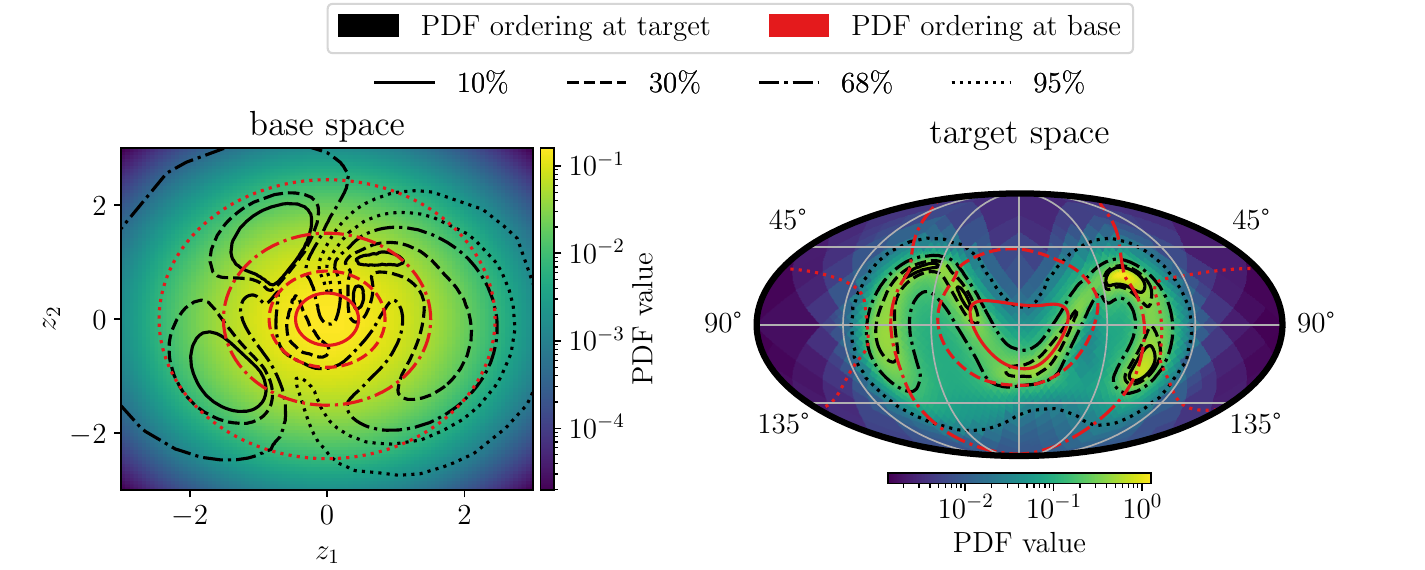}
    \caption{"M" shape on sphere - base-ordered contour is aligned with target-ordered contour}
    \end{subfigure}
    \caption{Illustration of base-ordered contours (red) and target-ordered contours (black) for a 1-d normalizing flow (a) and for a spherical normalizing flow (b)+(c) with different coverage probabilities. The spherical example illustrates that the base-ordered contour can be misaligned with the target-ordered contour. The spherical examples also show the inverse transformations of target-ordered contours in the base space.}
\label{fig:coverage_interval_transformation}
\end{figure*}
Both of these coverage calculations require the assumption of the large-data limit, but are very efficient because they do not require a large numerical effort. With normalizing flows that have a Gaussian distribution as base distribution, we can use a similar methodology to calculate coverage probabilities for contours from a target distribution of \emph{any} shape and dimension without resorting to numerical integration. The resulting coverage results are obtained for specific unique credible intervals that are defined by the base distribution. This is explained in the following. 

\begin{figure*}
    \centering
    \begin{subfigure}[t]{0.48\textwidth}
    \begin{tikzpicture}[baseline=-1.7cm]
    
	\node (normal) [circle, draw,inner sep=0pt, minimum size=2cm] {};
	
	

	\node (posterior_eq) [right of=normal, xshift=5cm] {$q_{\phi}(z_o)$};
	
	\node (base_dist) [left of=normal, xshift=-0.5cm] {$p_b(\hat{z})$};	
	
	\draw [] plot [smooth cycle] coordinates { ($(posterior_eq.west)+(-1.5,0.6)$) ($(posterior_eq.west)+(-1.65,0.45)$) ($(posterior_eq.west)+(-1.35,-0.3)$) ($(posterior_eq.west)+(-1.5,-0.9)$) ($(posterior_eq.west)+(-0.6,-0.6)$) ($(posterior_eq.west)+(-0.15,-0.6)$) ($(posterior_eq.west)+(-0.3,0.3)$) };
	
	\node (sample_left) [left of=normal, xshift=0.5cm, yshift=0.4cm] {$\hat{z}_i$};
	
	\node (sample_left_circ) [circle, draw, right of=sample_left, xshift=-0.7cm, inner sep=0pt] {};
	\node (sample_left_circ_shadow) [circle, right of=sample_left_circ, xshift=-0.9cm, inner sep=0pt] {};
	
	\node (chi2_node) [circle, above of=sample_left_circ, xshift=1.5cm, yshift=1.5cm, inner sep=0pt] {$\displaystyle -2\cdot \big( \mathrm{ln}p_b(\hat{z}_i)-\mathrm{ln}p_b(0)\big) \sim \chi^2$};
		
	\node (chi2_node_shadow) [circle, above of=chi2_node,yshift=-1.5cm] {};

	\node (sample_right) [left of=posterior_eq, xshift=-0.5cm] {$z_{o,i}$};
	\node (sample_right_circ) [circle, draw, right of=sample_right, xshift=-0.7cm, inner sep=0pt] {};
	
	\path[->] (sample_right) edge[bend right] node (rho_1) [below, xshift=-0.1cm] {$\rho^{-1}(z_{o,i})$} (sample_left_circ_shadow);
	
	\path[->] (sample_left) edge[bend left] (chi2_node_shadow);
	
	\end{tikzpicture}
	\caption{Illustration of $\chi^2$ behavior in the base distribution.} 
    \end{subfigure} 
    \centering
    \begin{subfigure}[t]{0.95\textwidth}
    	\includegraphics[width=\textwidth]{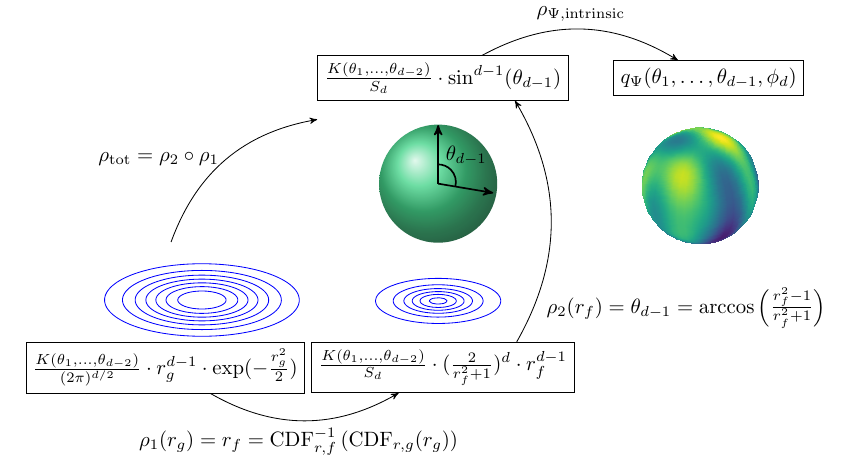}
    	\caption{Illustration of a stable normalizing flow $q_\Psi$ on the $d$-sphere starting with a $d$-dimensional standard normal distribution to allow for exact coverage calculation. The flows $\rho_1$ and $\rho_2$ involve only the radial coordinate due to rotational symmetry. The second flow $\rho_2$ is equal to the stereographic projection with a hyper plane that splits the sphere into two hemispheres. This is different from the visualization for illustrative purposes.} 
    \end{subfigure}

\caption{Coverage schematics to indicate the connection to the $\chi^2$ distribution at the base distribution (a) and to indicate the discussed mapping strategy to calculate coverage for spherical distributions (b).}
\label{fig:dsphere}
\end{figure*}

Figure \ref{fig:coverage_interval_transformation} a) illustrates that central simply connected\footnote{With "simply connected" we mean simply connected in the topological sense.} credible intervals at the base distribution transform to simply connected intervals in the target space because of the diffeomorphism connecting the two spaces. The normalizing-flow relation (eq. \ref{eq:nf_definition}) ensures that the interval in the target space still covers the same probability mass. This interval will be called "base-ordered" in the following, as it is the unique interval obtained using an ordering principle in the base PDF. If no ordering principle is applied, there are infinitely many ways to find an interval that contains a given probability mass. A similar issue arises in Frequentist confidence intervals, where a unique interval can be obtained by a likelihood ordering principle \cite{feldman_cousins}. Importantly, the base-ordered interval is always simply connected and in general different from the unique interval obtained by ordering the probability mass directly in the target space, as indicated in the right plot in Fig. \ref{fig:coverage_interval_transformation}. As such, for a given normalizing-flow there are always two unique credible intervals: a "base-ordered" interval that is transformed into the target space, and a "target-ordered" interval that is directly constructed in target space. 
The construction of the target-ordered interval requires numerical integration, in particular to calculate coverage probabilities for these intervals. In contrast, the base-ordered interval can be analytically calculated in base space utilizing the Gaussian base distribution, and transformed via the normalizing flow mapping interval to the target space. This in principle works for any dimension or target shape. More importantly, coverage probabilities can also be analytically calculated by utilizing a statistical relationship between the Gaussian distribution and the $\chi^2$ distribution.
This is indicated in Fig. \ref{fig:dsphere} a). If the samples $z_{o,i}$ follow the target $q_\phi(z_o)$, the corresponding samples $\hat{z}_i$ at the base must follow a standard normal distribution, and this implies the quantity $\lambda_{\mathrm{base}}=-2\cdot(\mathrm{ln}p_b(\hat{z}_i)-\mathrm{ln}p_b(0))$ is again $\chi^2$ distributed. This fact can be used to calculate coverage probabilities of the base-ordered contours, which by probability conservation (eq. \ref{eq:nf_definition}) is also valid for the transformed base-ordered intervals in target space.
The same idea works for distributions on manifolds like spheres. In \cite{flow_on_manifolds} the authors discuss how to define a flexible normalizing flow on a sphere via stereographic projection of a normalizing flow in the plane. A more stable alternative turns out to be directly to start with a base distribution on the sphere and parametrize a flexible flow which is intrinsic to the manifold \cite{Rezende2020}. We can combine both of these ideas to define a flexible flow on the sphere that also allows to define coverage. The methodology is illustrated in Fig. \ref{fig:dsphere} b) and consists of multiple sub-flows.

The base distribution is again a standard normal distribution in $\mathbb{R}^n$. It is transformed to another distribution in $\mathbb{R}^n$ that itself corresponds to the flat distribution on the $n$-sphere once it is stereographically projected. Then follows the stereographic projection, and finally the intrinsic flow on the sphere. The first two flows from the standard normal in $\mathbb{R}^n$ to the flat distribution on the sphere can be combined without actually invoking a Riemannian manifold flow which normally would involve more complicted Jacobian factors \footnote{Only an appropriate $\mathrm{sin}(\theta)$ factor is required for the log-determinant update.}(see eq. \ref{eq:manifold_density_update}).  To do so, the distribution in the plane is first written in spherical coordinates. The radial coordinate of the Gaussian distribution in $\mathbb{R}^n$ is then transformed to $\theta_{d-1}$ on the sphere, while the remaining angular coordinates $(\theta_1, \ldots, \theta_{d-2}, \phi)$ stay unchanged. The end result of these transformations turns out to be (see appendix \ref{appendix:spheres} for a derivation)

\begin{align}
\begin{split}
\rho_{\mathrm{tot},1} &= \biggl(\theta_1, \ldots, \theta_{d-2}, \phi, \pi\cdot(1-\mathrm{erf}(r_g/\sqrt{2})\biggr) \\
& \mathrm{(d \ = \ 1)} \label{eq:one_sphere_flow}
\end{split}
\\
\begin{split}
\rho_{\mathrm{tot},2} &= \biggl(\theta_1, \ldots, \theta_{d-2}, \phi, \\
&\mathrm{arccos}\left(1-2\cdot \mathrm{exp}(-r_g^2/2)\right)\biggr) \\ 
& \mathrm{(d \ = \ 2)}
\end{split}
\end{align}
for the special cases of the $1$-sphere and $2$-sphere, respectively. For higher-dimensional spheres, no analytical function for the radial part can be written down, but individual sub-parts of the flow are analytical. Evaluations and inverses can be obtained using bisection and potentially Newton iterations if derivatives are desired.
An issue that potentially arises with base-ordered credible intervals, in particular those defined on the sphere, is illustrated in Fig. \ref{fig:coverage_interval_transformation} b) and c). Because the number of possible transformations for a given flow function class is often large, there can be many different ways to generate similar target-ordered contours, which correspond to different relative base-ordered contour alignments. In Fig. \ref{fig:coverage_interval_transformation} b) a spherical normalizing flow based on the recursive flow in \cite{Rezende2020} is trained to describe a shape of the letter "M" on the sphere. After training, the $10 \%$ and $30 \%$ base-ordered contours enclose regions of low PDF values, and are misaligned with the respective target-ordered contours. In the extreme case they can even be on the opposite side of the sphere. The exact numerical amount of over- and under-coverage for the two contour types is in general always slightly different, but in such a situation, a coverage calculation that would indicate over-coverage with respect to the target-ordered contours can simultaneously indicate under-coverage with respect to the base-ordered contours. Only when coverage is exact, i.e. the contours enclose the true values exactly as predicted, the two contour definitions agree. In practice however, one often has slight over and under coverage, and one wants to ensure that both definitions at least agree on the sign and that their "center of gravity" overlaps. In order to achieve this, one can add an extra regularization term to the loss function during training. In Fig. \ref{fig:coverage_interval_transformation} c), the same normalizing flow is trained on the same target shape, but now with an added spherical contour regularization term $R_c$

\begin{align}
     R_c=\underbrace{-\mathbf{x}_{mean} \cdot \mathbf{x}_{\mathrm{base,mid}}}_{\mathrm{mean \ adjustment}} + \underbrace{\mathrm{var}\left(\{-\mathbf{x}_{mean} \cdot \mathbf{x}_{\mathrm{i, base, 50}}\}_{{}_N} \right)}_{\mathrm{centering \ mean \ in \ 50 \% \ contour}}  .
\end{align}
Here $\mathbf{x}_{mean} $ is the mean of the target PDF, $\mathbf{x}_{\mathrm{base,mid}}$ the base mean in target space, and $\mathbf{x}_{\mathrm{i, base, 50}}$ a point $i$ on the $50 \%$ base contour transformed to target space. The first term, a dot product of the target mean with the center of base-ordered contours, adjusts the center of the base-ordered contours to be aligned with the mean of the distribution. It avoids the situation where the base-ordered contour is aligned antipodal to the target-ordered contour. The second term, the variance of dot products of the target mean with $N$ points on the $50 \%$ base-ordered contour, adjusts the $50 \%$ base contour in target space to symmetrically surround the mean of the target distribution.
In principle, several regularization terms are possible, but the above terms work reasonably well. In Euclidean space, dot products would be replaced by Euclidean distances. If the PDF is highly degenerate and the target-ordered contours consist of disjoint sets, for example if the PDF consists of two narrow peaks that are widely apart, even regularized base-ordered contours might not be reliable in the sense of the same sign as the target-ordered contours - for certain probabilities base-ordered contours might indicate overcoverage while the corresponding target-ordered contours might indicate undercoverage, or vice versa. In these situations one either has to resort to numerical techniques to evaluate target-ordered contours, since those are usually the ones of interest, or directly work with the base-ordered contours. For visualization purposes, an issue might arise with base-ordered contours when the PDF has more than 2 dimensions. In this case one typically uses 1-d and 2-d marginal views of the PDF, and base-orderend marginal contours are not trivial to compute, while target-ordered contours can simply be constructed using samples, as is for example done in Fig. \ref{fig:gof_supplemantary} b) for a 4-d PDF. It is however possible to construct base-ordered contours for conditional sub-parts in an autoregressive PDF. It might be possible to create a resulting marginal base-ordered contour with the correct coverage via an appropriate combination of such conditional base-ordered contours, which could then be included in visualizations of marginal distributions. We leave this for future work.

\begin{figure*}
    \centering
    \begin{subfigure}[t]{0.48\textwidth}
    	\includegraphics[width=\textwidth]{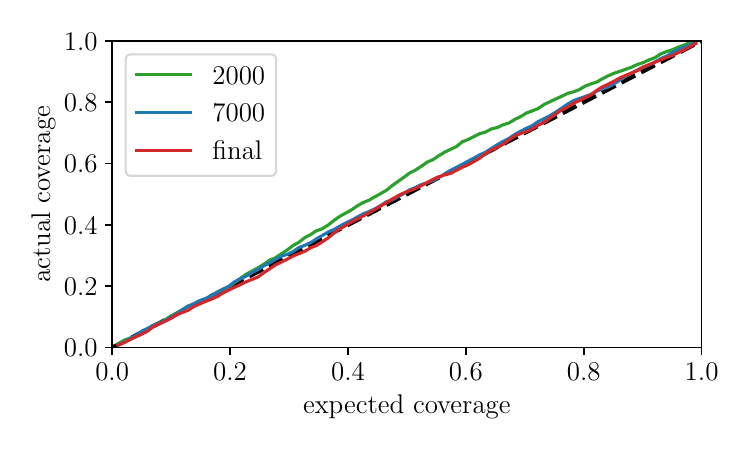}
    	\caption{Coverage curve (base-ordered)} 
    \end{subfigure}
    \begin{subfigure}[t]{0.48\textwidth}
    	\includegraphics[width=\textwidth]{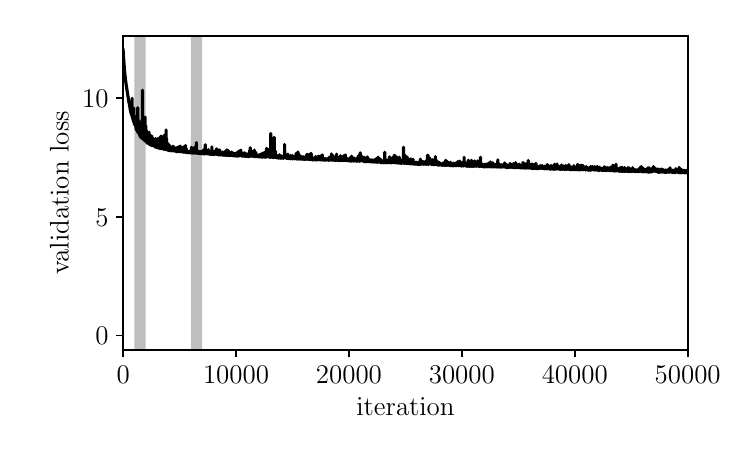}
    	\caption{Validation loss} 
    \end{subfigure}
 
    \caption{Illustration of the coverage behavior for a posterior calculated with dataset 3 (a). The coverage is calculated for different stages of training (iteration in the legend), including the end of training where the model is averaged using SWA. The validation loss curve is shown in (b), where the iterations in fig (a) are indicated by the shaded gray area.}
\label{fig:coverage}
\end{figure*}

We can test coverage in a coverage plot as indicated in Fig. \ref{fig:coverage} which shows expected versus actual coverage probabilities of base-ordered contours for a 3-d posterior over position and direction and the corresponding validation loss curve.
The posterior in this example is trained on dataset 3, in which the position, direction and energy of neutrino interactions are different from event to event. We use a factorized posterior $q_{\phi}(x_{p},y_{p},\theta;x)=q_{\phi_1}(x_{p},y_{p};x)\cdot q_{\phi_2}(\theta;x_{p},y_{p},x)$, which allows to learn the spherical part separately from the Euclidean part in a stable manner via an autoregressive structure \cite{iaf}. While the Euclidean part again uses a 2-d Gaussianization flow, the spherical part employs the previously described strategy to reach a flat distribution on the $1$-sphere (eq. \ref{eq:one_sphere_flow}) and afterwards uses convex Moebius transformations \cite{Rezende2020} parametrized by a neural network as an intrinsic flow. Coverage is calculated using the base-ordered contours. This is an example for coverage of a joint posterior defined on $\mathbb{R}^2 \times \mathcal{S}^1$, for which a numerical calculation of target-ordered coverage is already computationally too time-consuming. The base-ordered coverage calculation, however, would work analogously for higher dimensions and on more than two manifolds without a noticable computational increase. 
 
We can observe interesting coverage behavior during training, as depicted in Fig. \ref{fig:coverage}. Once the first phase of training is over, in which the loss function decreases rapidly, good coverage seems to be already reached, even though the overall training has not finished yet. The second phase, in which the optimization is much slower has been called "random diffusion phase" by \cite{tishby_ib}. The authors argue the network learns a better compression of the input data in this stage. In the context of normalizing flows, it seems the first phase brings the learned posteriors in line with the labels to reach proper coverage. In the second diffusion phase, the posterior regions shrink as much as they can while maintaining coverage. This is an empirical observation, and we leave more in-depth studies for future work.

\section{Systematic Uncertainties}
\label{sec:systematics}

\begin{figure*}
    \centering
    \begin{subfigure}[t]{0.48\textwidth}
    	\includegraphics[width=\textwidth]{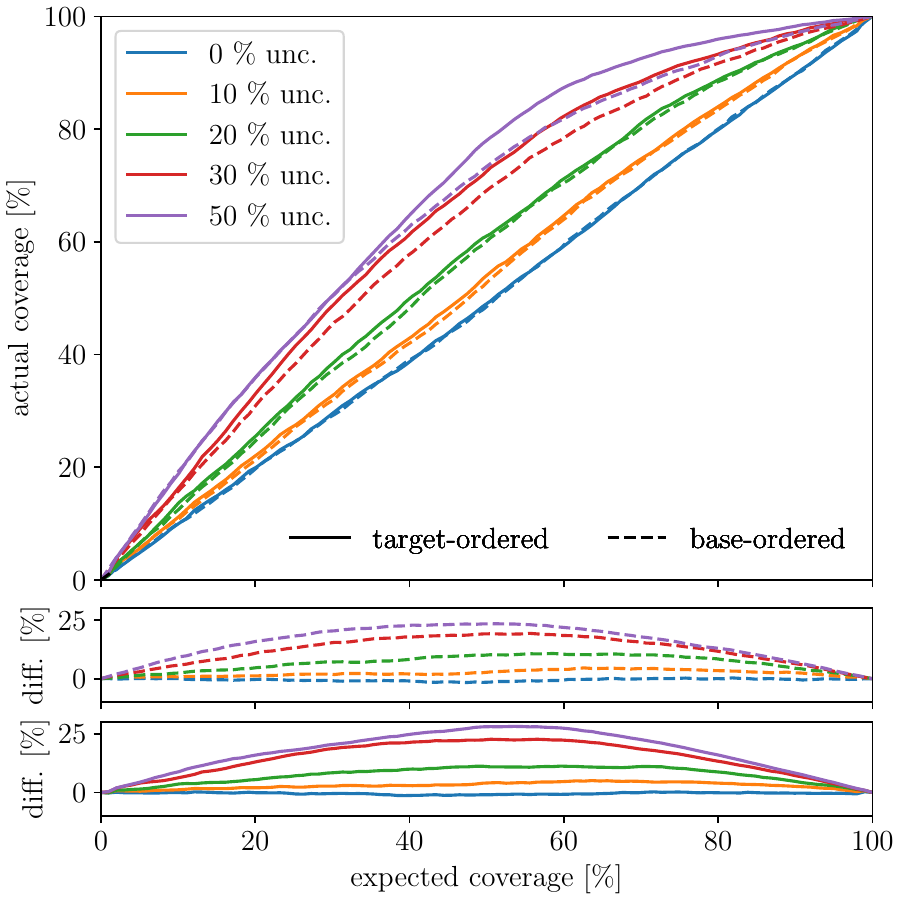}
    	\caption{Coverage curves for different systematics models.} 
    \end{subfigure}
    \begin{subfigure}[t]{0.48\textwidth}
    	\includegraphics[width=\textwidth]{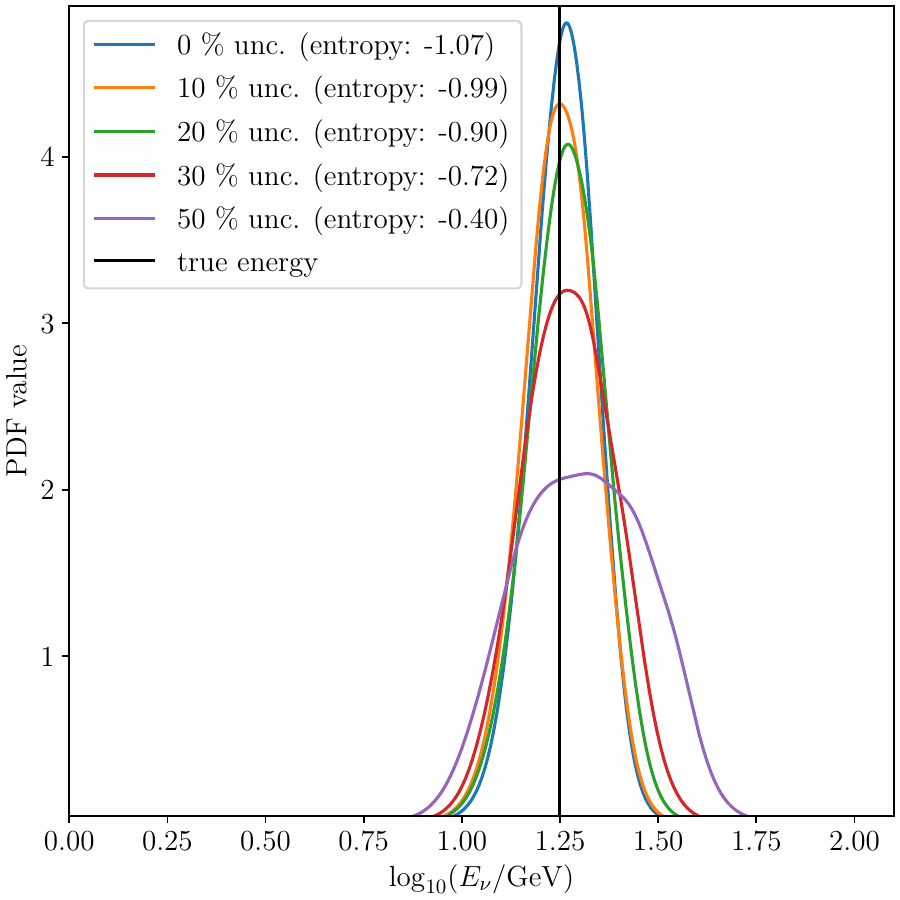}
    	\caption{Posterior distributions for different systematics models. The differential entropy is shown in brackets.}
     
    \end{subfigure}
    \caption{Illustration of the effect of systematic uncertainty inclusion via effective marginalization. Each model is trained on data with a different flat prior on the overall light yield per energy deposition (in percentage). The test dataset has no light yield uncertainty included. The left figure (a) shows actual coverage versus expected coverage for base-ordered and target-ordered contours using the test dataset. The right figure shows posterior distributions for an example event from the test dataset.}
\label{fig:coverage_with_systematics}
\end{figure*}

A standard practice to incorporate systematics in both Frequentist and Bayesian analyses is to first assume they can be parametrized by a parameter $\nu$ which follows a statistical distribution, for example a Gaussian with a known mean and width, e.g. $p(\nu;\mu_\nu,\sigma_\nu)$. Such a parametrization does not make sense in a Frequentist sense, but should rather be understood subjectively as the ignorance about the true value of the systematic parameter $\nu$ in question. In a Frequentist analysis, usually every systematic is then included as a log-probability penalty term in the likelihood function with respective nuisance parameters in a profile likelihood approach \cite{cranmer_2011}. In a Bayesian analysis the systematic uncertainties influence the joint distribution of $x$ and $\theta$, i.e. $p(x,\theta)\rightarrow p(x,\theta;\nu)$. A marginalized joint density can then be obtained as $\mathcal{P}_{t,M}(x,\theta)=\int p(x,\theta;\nu) p(\nu) d\nu$. Because it applies to the joint distribution, it also applies to any term in Bayes' theorem,

\begin{align}
\mathcal{P}_{t}(\theta;x,\nu)=\frac{\mathcal{P}_{t}(x;\theta,\nu)\mathcal{P}_{t}(\theta;\nu)}{\mathcal{P}_{t}(x;\nu)} ,
\end{align} 
and one can for example obtain the marginalized posterior as 

\begin{align}
\mathcal{P}_{t,M}(\theta;x)=\int \mathcal{P}_{t}(\theta;x,\nu)\cdot p(\nu) d\nu,
\end{align} which now includes the extra uncertainty about the systematics parameter $\nu$. 

In the previous sections we discussed how supervised learning and VAEs approximate the underlying distributions of interest via variational inference. A Monte Carlo simulation that first draws systematic parameters $\nu$ and then records events with labels $\theta$ and data $x$ according to the detector response will automatically produce samples from $\mathcal{P}_{t,M}(\theta,x)$ or any true conditional distribution of interest. These are the distributions that the neural network will approximate. Incorporating systematics is therefore possible at no change of the underlying procedure, as long as the simulation one is using for training additionally samples from the systematic distributions. No explicit form of the PDF of the systematic parameters has to be known, only sampling is required. For supervised learning, for example, one can just replace any term $\mathcal{P}_t$ by $\mathcal{P}_{t,M}$ in eq. \ref{eq:supervised_loss} and the neural network based approximation $q_\phi$ will approximate the marginalized true posterior $\mathcal{P}_{t,M}(\theta;x)$. It is already common practice in modern high-energy physics experiments to generate Monte Carlo simulations with potentially marginalized systematic parameters \cite{Aartsen2019}, so there is often no computational overhead in actually including those uncertainties.

To illustrate the effect of marginalization over systematic uncertainties, Fig. \ref{fig:coverage_with_systematics} shows actual versus expected coverage of the contours of a posterior over the logarithmic energy $\mathrm{log}_{10}(E_\nu/\mathrm{GeV})$.
The systematic uncertainty in this context is the expected number of observed photons given a certain energy deposition, i.e. the light yield, which is a common systematic uncertainty in neutrino detectors. In this example, it is modeled as a flat prior on the light yield with varying width with up to $50 \%$ relative uncertainty. For a given prior width, the normalizing flow is trained on dataset 4, which includes 4 degrees of freedom - position, direction and energy - and for each event the per-event light yield is additionally drawn with a random draw from the prior. As shown in Fig. \ref{fig:coverage_with_systematics} a), the wider the systematic prior that is used in the simulation the higher is the over-coverage of the true values. Fig. \ref{fig:coverage_with_systematics} b) shows the different PDFs for an example event from the test dataset. As expected, the PDFs get wider and have larger entropy for a larger width of the light-yield prior that is used during training.
In general, the result from base-ordered and target-ordered contours is pretty similar, which is reflected in the PDF shapes which are unimodal and pretty close to Gaussian. For $30 \%$ and $50 \%$ systematic uncertainty though, the PDFs become slightly non-Gaussian and the respective target- and base-ordered coverage probabilities start to diverge, as expected.

\section{Goodness-of-Fit of the neural network model}
\label{sec:goodness_of_fit}

\begin{figure*}
    \centering
    \begin{subfigure}[t]{0.5\textwidth}
        \centering
\begin{tikzpicture}[
title/.style={font=\fontsize{6}{6}\color{black!50}\ttfamily},
typetag/.style={rectangle, draw=black!50, font=\scriptsize\ttfamily, anchor=west},
node distance=1cm,>=latex']

	\node (normal) [circle, draw] {};
	
	
	
	\node (posterior_eq) [below of=normal_eq, yshift=-1cm] {$q_{\phi}(z_o;x)$};
	\node (outer_posterior) [fit=(posterior_eq), draw, rounded corners] {};
	\node (x) [circle, left of=posterior_shadow, xshift=-1.5cm] {$x$};
	
	\path[->] (normal) edge[bend left] node (phi_node) [left] {$\rho_{\phi}(\hat{z};x)$} (posterior_eq);
	\path[->] (x) edge[bend left] node (f) [left] {} (phi_node);
	\node (posterior_eq_u) [below of=posterior_eq, yshift=-0.1cm] {$q_{\varphi}(z_u;z_o,x)$};
	\node (posterior_shadow_u) [left of=posterior_eq_u,text width=0.8cm,circle] {};
	\node (outer_posterior_u) [fit=(posterior_eq_u), draw, rounded corners] {};

	\node (normal_input_u) [circle,draw, below left of=posterior_shadow_u, xshift=-1.5cm] {};

	\draw[->] (normal_input_u) edge[bend left] node (phi_node_u) [above,xshift=-0.3cm] {$\rho_{\varphi}(\hat{z};x,z_o)$} (posterior_eq_u);
	
	\path[->] (x) edge[bend left] node (f) [left] {} (phi_node_u);	
	
	\path[->] (posterior_eq) edge[bend right] node (f) [left] {} (phi_node_u);

	

	\node (prior_fit) [fit=(outer_posterior)(outer_posterior_u), draw, rounded corners] {};
	
	\node (prior_eq) [draw,rounded corners, right=of prior_fit,yshift=0.8cm, xshift=-0.8cm] {$p_{\Psi}(z_o,z_u)$};

\node (normal_for_prior) [draw, circle, above right=of prior_eq] {};
\path[->] (normal_for_prior) edge[bend right] node (f) [left, xshift=-0.1cm] {$\rho_\Psi(\hat{z})$} (prior_eq);

\node (llh_eq) [draw,rounded corners, right=of prior_eq,yshift=-1cm] {$p_{\theta}(x;z_o,z_u)$};

\node (normal_for_llh) [draw, circle, below left=of llh_eq] {};
\path[->] (normal_for_llh) edge[bend left] node (llh_rho) [left] {$\rho_{\theta}(\hat{z};z_o,z_u)$} (llh_eq);
\path[->] (prior_fit) edge[bend left] node (f) [left] {} (llh_rho);
\end{tikzpicture}

 \end{subfigure}
    \caption{Generic structure of the extended supervised loss for goodness-of-fit tests and semi-supervised applications based on four normalizing flows. The respective base space Gaussian random variables $\hat{z}$ are denoted by circles and the respective flow-defining functions by $\rho$. The arrows from the variational distributions $q$ to the flow-defining functions $\rho$ indicate sampling, where the samples serve as an input to $\rho$.}
    \label{fig:ext_supervised}
\end{figure*}

The coverage calculation is only a meaningful information for data that is similar to the training data. If new input data is given to a fully trained neural network, for example real data instead of a Monte Carlo simulation, it is desirable to  test if this new data can be described by the neural network model. If this is the case then the coverage results might apply\footnote{We presume coverage is calculated using the validation dataset and that validation and training dataset are sufficiently similar.}. Traditionally, a test between a model and data is done via a goodness-of-fit test. In the following we describe how the extended supervised training procedure described in Section \ref{sec:extended_supervised} allows to calculate a goodness-of-fit via posterior predictive checks. The structure of the extended supervised model is shown in Fig. \ref{fig:ext_supervised}. On top of a posterior $q_\phi$, we have an additional posterior $q_\varphi$ over unobserved latent variables, and a prior $p_\psi$ and $p_\theta$ as a generative model. All of those are (conditional) normalizing flows, and are crucial to calculate for posterior predictive simulations. Posterior predictive checks are a standard methodology in Bayesian analysis to do goodness-of-fit tests of the resulting posterior distribution (see \cite{bayesian_data_analysis2004}, chapter 6.3). In the extended supervised scheme, a posterior-predictive p-value, sometimes also called "Bayesian p-value"\cite{bayesian_data_analysis2004}, can be defined via

\begin{align}
p_{\textrm{val}}=\int\limits_{x,z} \bm{I}_{T(x,z)>T(x_{\mathrm{obs}}, z)} p_{\theta}(x;z) q_{\phi}(z;x_\mathrm{obs}) dx dz, \label{eq:pvalue}
\end{align}
and numerically calculated using samples from $p_{\theta}(x;z) \cdot q_{\phi}(z;x_\mathrm{obs})$, i.e. from the posterior predictive distribution. The quantities $T(x,z)$ are called test quantities, and $\bm{I}_{T(x,z)>T(x_{\mathrm{obs}}, z)}$ the indicator function, where the integral effectively counts the fraction of samples where $T(x_i,z_i)>T(x_{\mathrm{obs}},z_i)$ for an infinite number of samples $x_i,z_i$. A comparison of a given p-value with the ensemble of p-values from the training dataset gives a goodness-of-fit criterion. In contrast to Frequentist goodness-of-fit testing, the test quantities also depend on the parameter $z$, and the p-value distribution of the true model can be non-uniform and is usually more peaked around $0.5$ \cite{bayesian_data_analysis2004}. Only in certain limits, for example for infinitely precise posterior distributions, does the p-value distribution approach the uniform distribution \cite{bayesian_data_analysis2004}. In our case, we additionally do not expect to reach a uniform distribution as the neural network model will only yield an approximation of the true model, in particular since we do not have access to the exact data-generating function, but learn an approximation of it in tandem with the posterior distribution. We therefore determine the null hypothesis p-value distribution from the training data distribution. We propose to use a normalized version of the logarithm of the data-generating PDF (eq. \ref{eq:likelihood})

\begin{align}
\begin{split}
&\ \ \ \ T(x=(\overline{k},\overline{t}),z) \\
&=\sum\limits_{j=1}^{N}\frac{-\lambda_j(\theta) + k_j \cdot \mathrm{ln}(\lambda_j(\theta)) - \mathrm{ln}(k_j!)}{N} \\ 
&\ \ \ + \sum\limits_{i=1}^{N_\mathrm{tot}} \frac{\mathrm{ln}p_{\theta}(t_i;z)}{N_\mathrm{tot} }
\end{split}
\end{align}
as a test quantity for the Poisson process data. The division by the number of detection  modules $N$ and number of total observed photons $N_{\mathrm{tot}}$ in the whole detector makes different realizations of the data comparable\footnote{We also tested the default data-generating PDF as a test quantity. However, the resulting p-value distribution is not as informative for a goodness-of-fit criterion due to larger fluctuations between realizations.}. For other data-generating PDFs, or a test quantity based on the posterior distribution, this division would probably not be necessary. 
Any test quantity based on the data-generating PDF or posterior can readily be calculated since they are part of the extended supervised training and modeled by normalizing flows. If those normalizing flows are bad approximations of the true underlying PDFs, the test quantity will be less constraining. Any improvement in approximation of the normalizing flows is therefore desired for a better overall goodness-of-fit procedure.

\begin{figure*}
    \centering
    \begin{subfigure}[t]{0.95\textwidth}
    	\includegraphics[width=\textwidth]{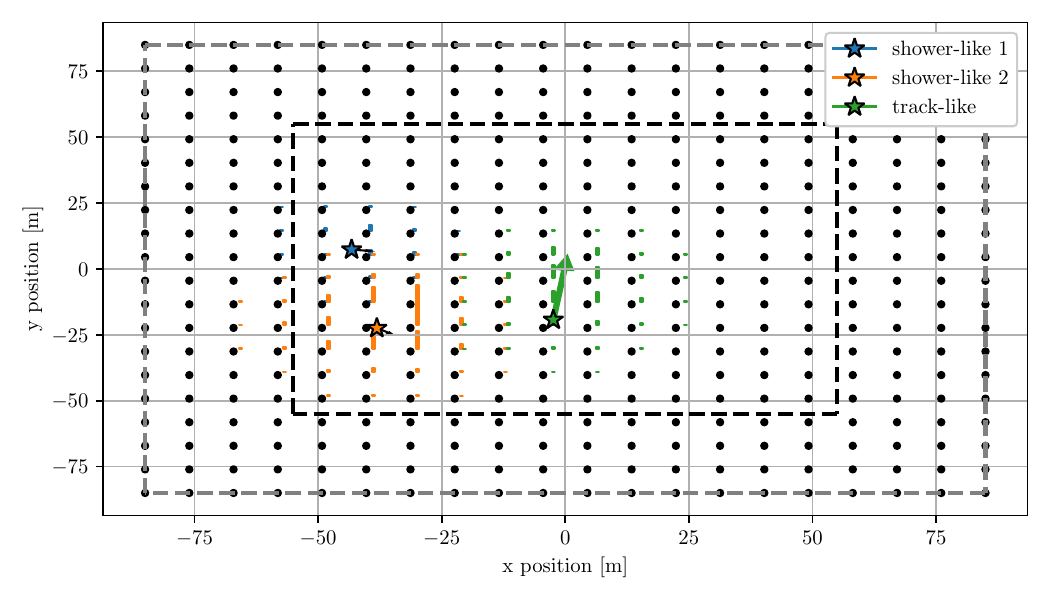}
    	\caption{Example interactions of three events overlaid on the detector used in datasets $5-7$.  Detector modules are denoted as black dots. The relative expected photon count in each module is visualized as vertical bars with the respective color.} 
    \end{subfigure}
    \\
    \begin{subfigure}[t]{0.95\textwidth}
    	\includegraphics[width=\textwidth]{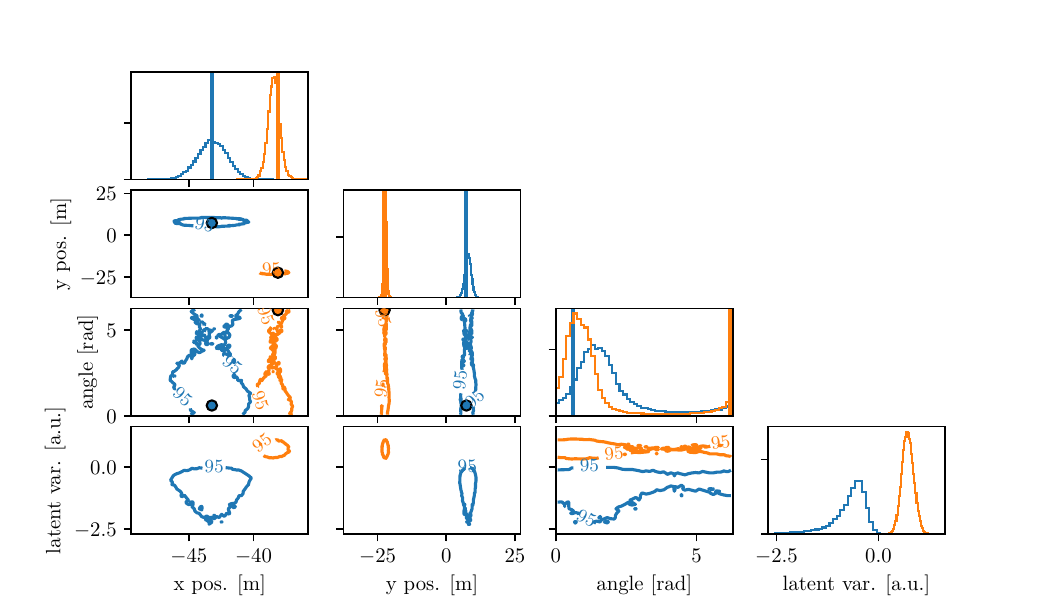}
    	\caption{Visualization of the final posterior approximation (marginal PDF in $1$-d and target-ordered contours of the marginal PDFs in $2$-d) for the two shower-like events visualized in Fig. \ref{fig:gof_supplemantary} a). The Monte Carlo truth is indicated with markers and vertical lines. }
    \end{subfigure}  
    \caption{A few events from the datasets for the goodness-of-fit example.}
\label{fig:gof_supplemantary}
\end{figure*}

In order to illustrate the overall procedure of goodness-of-fit testing, the extended supervised training is performed on dataset 5 (see table \ref{table:datasets} for information on all datasets) using the loss function described in eq. \ref{eq:extsupervised_loss}. In principle, the posterior $q_\phi$ could be a model trained in a purely supervised fashion as described in section \ref{sec:supervised}, with its weights fixed, and only afterwards the additional normalizing flows are trained using the loss function in eq. \ref{eq:extsupervised_loss}. However, in the following we train everything together from scratch, including the supervised term and the ELBO-like term. For the ELBO-like term, we sample one latent variable $z_{u,i}$ per observed input pair $z_{o,i}, x_i$ using the reparametrization trick, and minimize the overall resulting loss function, including $\mathcal{L}_{\mathrm{supervised}}$, with stochastic gradient descent. The training evolves similar to supervised training until the overall loss function does not reduce significantly anymore (see appendix \ref{appendix:impl_details} for training details). Dataset 5 contains shower-like interactions in a detector consisting of 400 modules, where the interactions are constrained to lie within a certain containment region (the black dashed region in Fig. \ref{fig:gof_supplemantary}). The photon arrival time PDFs $p(t;z_o,\textbf{x})$ that appear in the data-generating function (eq. \ref{eq:likelihood}) are modeled by a 1-dimensional conditional normalizing flow that takes as extra information the position of each module. Additionally, the mean count expectations of the Poisson distributions are predicted, and together with the normalizing flow, allow to describe the complete data-generating function. 
In the following, we apply this trained model also to events of dataset 6 (uncontained showers) and dataset 7 (track-like events). Figure \ref{fig:gof_supplemantary} a) shows the interaction of three example events involved in dataset 5 and 7. The first two events are a low energy (blue) and high-energy (orange) shower event from dataset $\mathrm{5}$. The third event is a $20 \ \mathrm{m}$ track-like event (green) from dataset $\mathrm{7}$ (see table \ref{table:datasets} for a list of all datasets). All three events are within the containment volume, as defined by dataset $\mathrm{5}$ (contained showers), which is also indicated as the black bounding box. The expected value of photon counts per module is shown as vertical bars next to each module. Figure \ref{fig:gof_supplemantary} b) shows a comparison of the final posterior distributions of the two shower events over the four degrees of freedom after training, .i.e. a visualization of the learned joint distribution over $z_o$ and $z_u$,  $q_\phi(z_o;x)\cdot q_\varphi(z_u;z_o,x)$. The energy is not used as a label during training, so the latent variable from the unsupervised part of the extended supervised training can be identified with the energy degree of freedom. The high-energy shower (orange) produces more photons, and allows to better inform about the true event properties than the lower energy shower.

\begin{figure*}[htpb]
    \centering
    \begin{subfigure}[t]{0.7\textwidth}
    	\includegraphics[width=\textwidth]{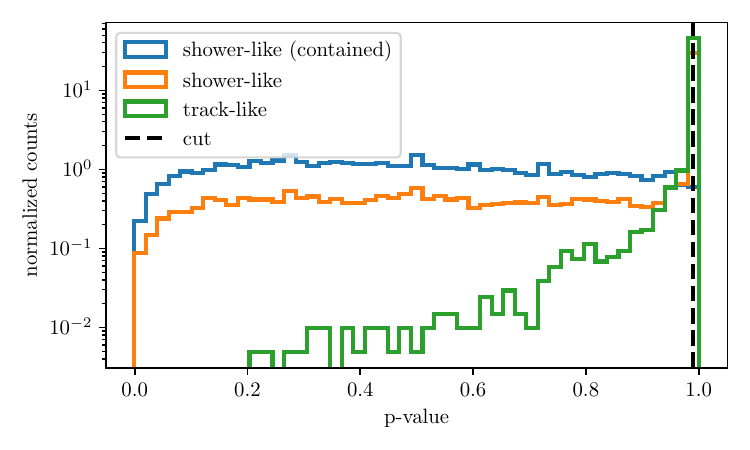}
 
    \end{subfigure}
   
    \caption{Goodness-of-fit p-value distribution for three datasets. The model is trained on a dataset which consists of contained shower-like neutrino interactions (dataset 5 - orange). The p-values are also calculated for a dataset of shower-like interactions in the whole detector (dataset 6 - blue) and a dataset consisting of $20 \ \mathrm{m}$ long tracks of light emission (dataset 7 - green) to emulate muon tracks.}
\label{fig:goodness_of_fit}
\end{figure*}

The p-value distribution of the test set of dataset 5 after training, calculated using eq. \ref{eq:pvalue}, is shown in Fig. \ref{fig:goodness_of_fit}
\begin{figure*}[htpb]
    \centering
    \begin{subfigure}[t]{0.99\textwidth}
    	\includegraphics[width=\textwidth]{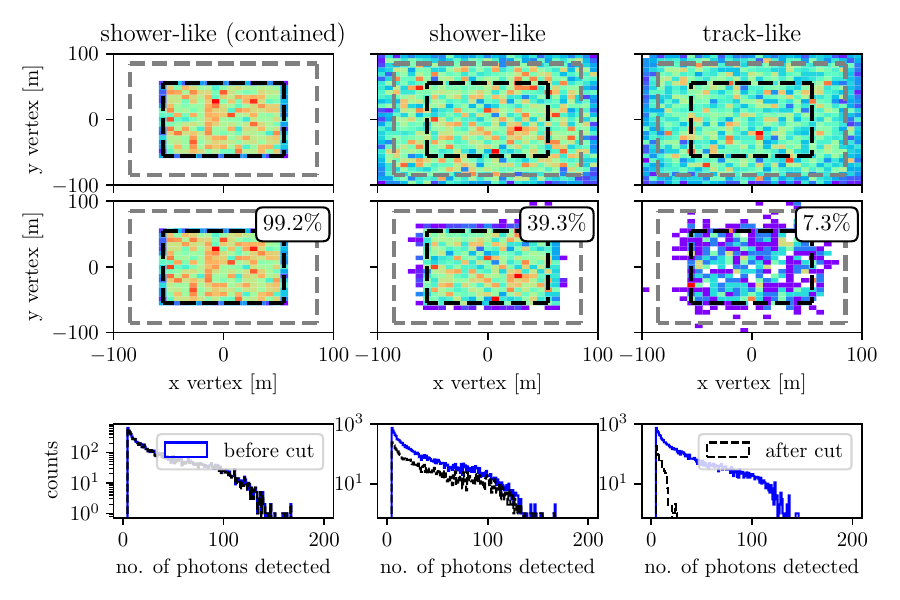}
 
    \end{subfigure}
   
    \caption{Effect of the p-value cut ($p_{\mathrm{val}} < 0.99$) on the spatial distribution (upper row vs. center row) and the distribution of the number of detected photons (lower row) for datasets $4$ to $6$. The center row also indicates the percentage of retained events for the respective dataset.}
\label{fig:goddness_of_fit_cut_effects}
\end{figure*}
as the orange curve. In addition, two other p-value distributions are depicted, where the final trained model is applied to events from dataset 6 (shower-like uncontained) and dataset 7 (track-like). As can be seen, the p-value distribution for dataset 5 is not totally uniform, possibly because the individual PDFs have not all fully converged during training. In general, full convergence of all normalizing flows is indicated when the p-value distribution of the training dataset is fairly symmetric. Towards the beginning of the training process, the p-value distribution of the training dataset is still highly asymmetric. The p-value distribution of dataset 6, uncontained shower events, has a similar shape as contained showers, except it also contains a peak for p-values close to $1$. The events that pile up close to $1$ are events that lie outside the containment region, which are not part of the training dataset and therefore get such a bad goodness-of-fit. For dataset $7$, which contains track-like events, there is a skew in the distribution which extends over the whole p-value range and most events have a p-value close to $1$. Figure \ref{fig:goddness_of_fit_cut_effects} shows the effect of applying a cut that retains events with a p-value smaller than $0.99$. The training dataset is almost unchanged, while the cut removes most events outside the containment region for uncontained shower-like events. For track-like events, additionally events inside the detector are substantially reduced. The distribution of the number of photons per event in the lower row indicate that the events that remain for the track-like dataset are low energy events that have an intrinsically low information content and are with the given model not statistically separable to a low energy shower-like event. Therefore, a simple energy cut can remove those events and one would end up with a strong separation, in this example a complete background reduction above a certain energy threshold. An alternative is a cut on high-entropy events, which is possible with normalizing flows, since they allow to calculate the per-event differential entropy to arbitrary precision. This can also be seen in Fig. \ref{fig:gof_supplemantary} b). The higher energy shower has a visibly lower overall entropy.
This example is obtained with toy simulations that only qualitatively model a situation commonly encountered in neutrino telescopes, in particular with a 2-d compared to a 3-d setting. However, with 400 detection modules and the inclusion of relevant effects like scattering, absorption and typical Cherenkov emission features, the data should broadly mimic a real world experiment like IceCube DeepCore \cite{deepcore} both in statistical and computational aspects. The result illustrates that it is in principle possible to train an extended supervised model on a signal dataset, apply the resulting model to any data event, and based on the resulting p-value decide to keep the event or exclude it from further analysis. The remaining events should be similar to events from the training sample, within uncertainties given by the training statistics and any included systematics. A further second cut on an energy variable or the entropy should then be enough to yield a statistically optimal event selection for the given model. If the machine learning model is improved, for example by a better data encoding scheme or a more powerful normalizing flow, the p-value separation and thereby the final event selection should also improve. Besides such an application for data selection, it might also be interesting to study goodness-of-fit results for concrete data to Monte Carlo comparisons or for data anomaly detection.

\section{Related Work}
\label{sec:related}

We discussed that standard supervised training can be viewed as variational inference, and variational inference can be viewed as an approximate likelihood-free inference approach.
An active topic in likelihood-free inference are neural sequential inference procedures \cite{sequential_neural}. These procedures typically require guided re-simulations for a single event, and eventually should converge to the true posterior. From our perspective, simple supervised learning with normalizing flows, in combination with coverage and goodness-of-fit guarantees as described earlier, might be more suited for certain applications in high-energy physics or astro-particle physics where processing time is valuable - for example when posteriors for large amounts of data are desired. Also single astronomical multi-messenger alerts \cite{icrc_multimessenger} can benefit. An initial posterior that can be trusted due to the discussed coverage and goodness-of-fit checks could be send to the community quickly. One can later always decide to refine the result with an iterative likelihood-free procedure like automatic posterior transformation (APT)  \cite{automatic_posterior_transformation} which would naturally blend in due to its use of normalizing flows.

For gravitational waves, posterior inference with autoregressive flows has been discussed recently in \cite{Green2020}. In this line of work, coverage is checked with 1-dimensional marginal distributions which seems more computationally demanding than our approach. Systematics or goodness-of-fit tests are not discussed in that paper.

Another related line of work discusses inference of physics posteriors specifically based on coupling layers \cite{cinns_2019}, originally dubbed conditional invertible neural networks (c-INNs), more recently using a conditional variant \cite{cinns_2021} of GLOW \cite{glow} where conditionality is added to the coupling layer inputs. For Euclidean normalizing flows in low dimensions, which is what we are interested in, there exist expressive normalizing flows without coupling layers like Gaussianization flows \cite{Meng2020}. These normalizing flows, similar to certain manifold flows, can be made conditional by predicting their whole parameter set by a neural network, which is what we use in this paper. A further difference to c-INNs is that we interpret the base space not as a latent space, but purely as an auxiliary space that can be utilized for coverage calculations.

Normalizing flows have also been used to model the likelihood function directly \cite{Wong2020}. Pure likelihood modeling inherits some problems of the standard likelihood machinery: a time consuming optimization or MCMC procedure. However, in our view it is useful to model the likelihood as the data generating PDF in the proposed "extended supervised training" (Section \ref{sec:extended_supervised}) in combination with an associated posterior. 

A central part of our derivation is based on the joint KL-divergence. The importance of the joint KL-divergence has been emphasized in other approximate Bayesian inference, for example to unify expectation propagation in Gaussian processes \cite{gp_joint} and as mentioned earlier for the derivation of the supervised loss for neural networks \cite{Martens2014}. In the latter example, however, the loss function is not called the log-posterior, but conditional maximum likelihood objective with respect to the neural network parameters. While both views are correct, the posterior interpretation is more fruitful in our opinion, in particular in combination with normalizing flows which shows that supervised learning can behave as rather well-approximated variational inference. Another related recent work applied normalizing flows for a neutrino oscillation analysis \cite{Pina-Otey2020}. The authors do not use the joint KL-divergence in the derivation, which makes it less obvious how to include systematics, although they briefly mention one approach among two alternatives that is similar to our suggestion. The authors also argue for a vague agreement of the base distribution with a Gaussian after training, but do not discuss quantifiable coverage tests.

Likelihood approaches typically only allow to define a rigorous goodness-of-fit for special random variables. However, realistic applications typically involve unbinned likelihoods, for which no rigorous goodness-of-fit can be defined \cite{gof_pitfalls}. Furthermore, coverage and systematics checks are often computationally challenging. For traditional machine learning based regression, and even for other methods with normalizing flows, as pointed out above, coverage or goodness-of-fit are either impossible or not straightforward to compute.

The proposed extended supervised training with conditional normalizing fĺows combines 1) systematics 2) coverage and 3) a rigorous goodness-of-fit measure in one variational inference algorithm, without restrictions to the underlying PDFs. We are not aware of any other Ansatz in the literature that has these properties.

\section{Conclusion}
\label{sec:conclusion}

In the first part of the paper we have shown that the supervised learning loss, an extended supervised loss and the unsupervised ELBO-loss of variational autoencoders can all be derived under one unifying theme of variational inference. The inference in all these approaches can be traced back to the minimization of the KL-divergence between the true joint distribution of data and labels and an approximation for certain sub-parts of the joint distribution based on neural networks. 

For supervised learning, the derivation of the loss function from the KL-divergence of the joint distribution is well-known but the connection to variational inference of the posterior distribution is typically not being made. From our perspective, however, this interpretation is very useful in high-energy physics, even when using an MSE loss function which corresponds to a standard normal posterior approximation. After all, a central task of a physics analysis is to calculate a posterior distribution, and the variational perspective shows that neural networks learn to predict exactly that. Conditional amortized normalizing flows with a standard normal base naturally generalize the MSE loss, and are crucial to make the variational inference perspective usable in practice. 

A simple replacement of observed with unobserved labels in the KL-divergence allows to eventually derive the ELBO loss of variational autoencoders. In this derivation, the usual encoder $q_\phi$ and decoder $p_\theta$ are both manifestly auxiliary, and are explicitly separated from the true distribution $\mathcal{P}_t$ which $p_\theta$ and $q_\phi$ both try to approximate within a given identifiability class of transformations. This is different from the usual derivation via the marginal likelihood, e.g. as discussed in the original VAE paper \cite{kingma2013}, where $p_\theta$ is often used to denote both the true posterior and the recognition model, which obfuscates the fact that those are typically not the same and also do not have to be the same parametric class.

Conditional normalizing flows with a Gaussian base distribution have a second advantage. They can be used to calculate coverage without numerical integration for certain unique "base-ordered" contours of the normalizing-flow PDF which are defined via the base space. In contrast to "target-ordered" contours created in the target space by starting at the highest PDF values, which can consist of disjoint sets, the base-ordered contours are always simply connected in the topological sense. We further expanded the idea to calculate coverage for distributions on m-spheres $\mathcal{S}^m$, or more generally to posteriors defined autoregressively on tensor products like $\mathbb{R}^n \times \mathcal{S}^m$. To enable coverage for such distributions we derived the transformation from the m-dimensional standard normal Gaussian to the flat distribution on the m-sphere, which for 1 and 2 dimensions is conveniently simple and invertible. A further intrinsic flow on the sphere can then describe a more complex distribution. This two-step procedure ensures stable training, since the transformation from the plane to the sphere is fixed and does not change during optimization. We also discussed a regularization term that can be added to the loss function in order to ensure that base-ordered and target-ordered contours align. This can be in particular important for spherical distributions. If the distributions are very non-Gaussian with widely separated peaks, the coverage probabilities for base-ordered and target-ordered contours can diverge - which is only a problem when over- or undercoverage is observed. In such a case extra care in the interpretation is necessary.

We demonstrate a coverage calculation for a joint 3-d posterior of position and direction ($x,y,\theta$) for shower-like neutrino interactions in a 2-d toy Monte Carlo. Empirically, we observe that coverage of the base-ordered contours is already obtained once the training phase enters the random diffusion phase as defined by \cite{tishby_ib}. Since the loss still decreases during this phase, only more slowly, and in the variational interpretation this implies shrinking posteriors, in this second training phase the posterior approximations get slowly better while coverage more or less stays intact.

Viewing all these models as coming from the joint KL-divergence makes it also natural to include systematic uncertainties in the final approximated posterior distributions. They can be included via sampling from systematics priors during the Monte Carlo generation process, which lets the normalizing flow approximate the marginalized true posterior distribution - a process one might call "effective marginalization". The training process of the model otherwise stays the same. We test this method with a dataset with a varying energy-scale systematic and an approximation to the energy posterior distribution. The resulting posterior regions are wider and are over-covering more for a higher systematic uncertainty, as expected.

Finally, we test the \emph{extended supervised learning} scheme. It is related to semi-supervised learning and involves a hybrid of the supervised and VAE loss functions. In addition to the label posterior $q_\varphi$, a latent posterior $q_\phi$, a prior $p_\psi$, and a decoder $p_\theta$ are learned either jointly or separately as an add-on to the supervised posterior $q_\varphi$. The extra distribution $q_\phi$ learns latent variables to describe information not captured by the posterior of known labels $q_\phi$, while the prior and decoder define a generative model. The combination of a posterior and a generative model allows to perform Bayesian predictive simulations, which in combination with a test quantity $T(x,z)$, which we suggest to be the normalized logarithm of the learned data-generating PDF, allows to calculate a p-value which can be used as a goodness-of-fit criterion. Any data that are not similar to the training data within the given uncertainty of the statistical approximations of the neural networks lead to p-values close to 1 and can be excluded. We test this concept with a neural network model trained on shower-like neutrino interactions contained in the center of a hypothetical detector consisting of 400 detection modules. In comparison with shower-like interactions in the full detector, the model calculates a small p-value for events on the boundary of the detector. We also apply it on a dataset containing track-like energy depositions mimicking muon tracks, for which the separation is  markedly visible. Such a goodness-of-fit selection strategy can potentially allow a very simple but effective data selection.

Let us mention two aspects that require further investigation. The first is related to the stability of the training algorithm and the final learned distributions. We used stochastic weight averaging to average over several iterations at the end of training to obtain a more stable posterior solution. This serves as an approximation to ensemble averaging, which is basically averaging under a variational posterior over network weights. Due to varying learning rates for different model complexities, slight fluctuations in the final loss values remained (see Fig. \ref{fig:posterior_vs_complexity}). Ideally, one would like a scheme where repeated training schedules lead to repeatable posterior regions which are indistinguishable within a certain margin which can be defined before training. This needs to be studied more systematically.

A second aspect is related to the data encoding.
In all studies we have split the conditional posterior into a data encoding based on a GRU and aggregate MLP, and a second MLP which further maps that encoding summary to the normalizing-flow parameters. Our Monte Carlo tests show that at some level of flow complexity the performance bottleneck of the posterior approximation can in part stem from the data-encoding architecture. This particular encoding was only chosen for simplicity and serves as a proof-of-concept, but it is not an optimal encoding. In an application of the method, it is advised to use the best encoding mechanism available for the given experiment. Any improvement in data encoding not only translates to tighter posterior contours, but also to a more powerful goodness-of-fit calculation.

\section*{Acknowledgements}
We thank Alexander Trettin for suggesting a more realistic parametrization for the toy Monte Carlo.


\section*{Code Availability}
In the course of this work a software package has been developed to handle the described product manifold normalizing flows and allow for straightforward coverage calculations. It is available as an open-source software package at \url{https://github.com/thoglu/jammy\_flows}.

\appendix


\section{Implementation details}
\label{appendix:impl_details}

\subsection{Training}

All training procedures use ADAM \cite{adam}, which gives much faster convergence than standard gradient descent (SGD) \cite{sgd}. For the experiments in section \ref{sec:example_gaussianization_flows} we use a mostly fixed learning rate which adjusts itself to keep fluctuations of the validation loss at a certain level. For the affine flows, we had to fix the learning rate completely, because too large learning rates would sometimes lead to drastic jumps in the loss function. This might be a problem that only occurs for flows which are defined by less than a handful of parameters. For Gaussianization flows  \cite{Meng2020}, similar behavior could be seen for single-layer flows with a single basis function. Once the number of flow parameters is increased beyond a certain level, it does not seem to be an issue anymore. For the experiments in section \ref{sec:coverage}, \ref{sec:systematics} and \ref{sec:goodness_of_fit} we start with a high learning rate and then use a decreasing learning rate scheduler that lowers the learning rate by fixed factors until it reaches a value of about $0.0001$. Towards the end of training, once the loss reaches a more or less constant level, the final iterations are averaged using stochastic weight averaging (SWA) \cite{swa}. As argued by the authors, SWA approximates model ensemble averages. Model averages, in turn, can be used to calculate the expectation value under the approximate weight posterior since SGD iterations itself can be interpreted as samples from a variational approximation of the weight posterior \cite{sgd_as_variational}.

\subsection{Posterior}

The posterior parametrization is indicated in Fig. \ref{fig:nf_illustration} (c). The first encoding part is similar for all experiments. A GRU encodes the input data sequentially, the output is aggregated and mapped by a single layer MLP to a representation $h$ that summarizes the data.

The second step in the posterior is another MLP that maps the summary representation $h$ to the respective flow parameters. We use two inner layers for this second MLP, and vary its number of dimensions for the various experiments. The parameter choices are described in Section \ref{appendix:parameter_choices}.

\subsection{Data-generating PDF in goodness-of-fit}

For the data-generating PDF for the extended supervised model used for the goodness-of-fit, we have a discrete Poisson part and a continuous part. We use a normalizing flow for the continuous part $p(t;z_o)$ of the generating distribution (eq. \ref{eq:likelihood}). The involved MLP that is used in the amortization also outputs a logarithmic mean expectation for photon counts to calculate the related Poisson distribution, so we model the discrete and continuous part in a joint prediction step. By simultaneously predicting the flow parameters and the Poisson mean, it is possible to model the correlation between the two. The input to the MLP are the event parameter labels $z_o$ and the module position in (x,y) coordinates.

\subsection{Parameter choices}
\label{appendix:parameter_choices}
 We choose a single-layer GRU with hidden dimension $10$ and for the aggregation MLP choose intermediate dimension $15$ and map to a $20$-dimensional representation $h$ for the first 3 datasets in the paper. This encoding part then has roughly $1000$ parameters. For the larger detector used in datasets 5-7 the involved dimensions are a bit larger. For the comparison in Fig. \ref{fig:posterior_vs_complexity} the dimensions in the two inner layers of the second MLP vary between $1$ and $100$. For the Gaussianization flow, additionally the number of flow layers varies between $1$ and $5$, and the number of basis elements per flow varies between $1$ and $10$, as the flow complexity increases. 

\section{Transforming a Gaussian to the flat distribution on the sphere}
\label{appendix:spheres}
The following calculation derives the transformation $\rho_{\mathrm{tot}}$  from the d-dimensional Gaussian distribution to the flat distribution on the d-sphere (see Fig. \ref{fig:dsphere} b)) which is discussed in Section \ref{sec:coverage}. The flow can be split up as $\rho_{\mathrm{tot}}=\rho_2 \circ \rho_1$.
The first flow $\rho_1$ transforms the standard normal to a distribution $p_f$ that corresponds to the distribution in the plane that is the stereographic projection of the flat distribution on the sphere. We can derive $p_f$ by starting with the flat distribution on the sphere. The flat distribution on the d-sphere is defined as \cite{hypersphere_flat}

\begin{align}
\begin{split}
p_{\mathrm{flat,sphere}}&=\frac{\mathrm{sin}(\theta_1)^{1} \cdot \ldots \cdot \mathrm{sin}(\theta_{d-1})^{d-1}}{S_{d}} \\
&= \frac{K(\theta_1,\ldots, \theta_{d-2}) \cdot \mathrm{sin}(\theta_{d-1})^{d-1}}{S_d},
\end{split}
\end{align}
where the sine factors start to appear at dimension two. The factor $K$ abbreviates the first $d-2$ of those factors and $S_d$ denotes the surface area of the d-sphere. Note that $\theta_1$ to $\theta_{d-1}$ take values between $0$ and $\pi$, and an additional angle $\phi$ takes values values between $0$ and $2\pi$. We then define a flow $\bm{\rho_{2}}^{-1}=\left(\theta_1, \ldots, \theta_{d-2}, \phi, \rho_{2}^{-1}(\theta_{d-1})\right)$ which is similar to a stereographic projection to the plane and which just transforms the last spherical coordinate $\theta_{d-1}$, while all other angles are left unchanged. The angle $\theta_{d-1}$ is always related to the $d$th embedding space coordinate $x_d$ via $x_{d}=\mathrm{cos}(\theta_{d-1})$ \footnote{Note that for $d=1$ the angle $\phi$, which normally is defined with respect to $(1,0)$ and goes from $0$ to $2\pi$, has to be redefined as $\theta_0=\phi+\frac{pi}{2}$ in order for this relation to hold. The angle $\theta_0$ further goes from $0$ to $\pi$, similar to the other $\theta_{d-1}$, and is measured with respect to the "north pole" of the sphere.}. In 3-dimensional space, for example, $x_d$ equals the $z$ coordinate. At the same time, for a stereographic projection onto a plane which aligns to split the sphere and has plane coordinates $\bm{x_p}$, it is known \cite{alg_hypersurfaces} that the relation of the  embedding coordinate $x_d$ to the plane coordinates is given by $x_{d}=\frac{(\sum x_{p,j})^2-1}{(\sum x_{p,j})^2+1}=\frac{r_f^2-1}{r_f^2+1}$, which is expressed here entirely as a connection to the radial coordinate in the plane via $r_f=\sum x_{p,j}$. It therefore follows that 

\begin{align}
\theta_{d-1}=\mathrm{arccos}\left(\frac{r_f^2-1}{r_f^2+1}\right)\equiv \rho_2(r_f)
\end{align}
or vice versa

\begin{align}
r_f=\sqrt{\frac{2}{1-\mathrm{cos}(\theta_{d-1})}-1}\equiv \rho_2^{-1}(\theta_{d-1}),
\end{align}
which is a relation that is indicated in Fig. \ref{fig:dsphere} b). Applying the flow $\bm{\rho_{2}}^{-1}$ to the flat distribution on the sphere we obtain

\begin{align}
\begin{split}
&\ \ \ \ p_f(\theta_1, \ldots, \theta_{d-2}, \phi, r_f) \\
&=p_{\mathrm{flat,sphere}}(\theta_1, \ldots, \theta_{d-2},\phi, \rho_2(r_f)) \cdot \left|\frac{d \rho_2(r_f)}{d \theta_{r_f}} \right|
\end{split} 
\\
\begin{split}
&=\frac{K(\theta_1,\ldots, \theta_{d-2})}{S_d} \cdot \left(1-\left(\frac{r_f^2-1}{r_f^2+1}\right)^2\right)^{(d-1)/2} \\
& \ \ \ \cdot \frac{2}{r_f^2+1} \cdot r_f^{d-1} 
\end{split}
\\
&=\frac{K(\theta_1,\ldots, \theta_{d-2})}{S_d} \cdot \left(\frac{2}{r_f^2+1}\right)^d  \cdot r_f^{d-1},
\end{align}
where $p_f$ is now the corresponding PDF in the plane after the stereographic projection. We now need to find the flow $\rho_1$ (indicated in Fig. \ref{fig:dsphere} b)) to transform a standard normal Gaussian distribution to $p_f$ or vice versa. Similar to the transformation from the sphere to plane, this transformation can be done entirely in the radial coordinate when the Gaussian distribution is written in spherical coordinates. The radial transformation can be calculated using the cumulative distribution functions of the radial PDFs.
The two radial PDFs are

\begin{align}
\begin{split}
p_{r,g}(r_g)&=\int_{\theta_1,\ldots,\theta_{d-2},\phi} \frac{K(\theta_1,\ldots,\theta_{d-2})}{(2\pi)^{d/2}} \cdot r_g^{d-1} \\ 
&\ \ \ \cdot \mathrm{exp}\left(-\frac{r_g^2}{2}\right) d\theta_1 \ldots d\theta_{d-2} d\phi \\
&=\frac{S_{d-1}}{(2\pi)^{d/2}}r_g^{d-1} \cdot \mathrm{exp}\left(-\frac{r_g^2}{2}\right)
\end{split}
\end{align}
and

\begin{align}
\begin{split}
p_{r,f}(r_f)&=\int_{\theta_1,\ldots,\theta_{d-2},\phi} \frac{K(\theta_1,\ldots, \theta_{d-2})}{S_d} \cdot \left(\frac{2}{r_f^2+1}\right)^d \\ 
&\ \ \ \cdot r_f^{d-1}    d\theta_1 \ldots d\theta_{d-2} d\phi \\
&=\frac{S_{d-1}}{S_d} \cdot \left(\frac{2}{r_f^2+1}\right)^d \cdot r_f^{d-1},
\end{split}
\end{align}
where $p_{r,g}$ is the radial distribution of the d-dimensional standard normal distribution, also known as the $\chi$-distribution, and $p_{r,f}$ the radial distribution of $p_{f}$. The corresponding CDFs which map from $\mathbb{R}^+$ to $[0,1]$ then follow to be \footnote{The general formula is evaluated using \emph{Wolfram Alpha} at \texttt{http://www.wolfram-alpha.com}. The calculation of $\mathrm{CDF}_{r,g}$ likely involves variable substitution, integration by parts, and then an identification with the upper incomplete gamma function. The calculation of $\mathrm{CDF}_{r,f}$ likely involves variable substitution and an iterative application of hypgergeometric identities.}

\begin{align}
\begin{split}
\mathrm{CDF}_{r,g}&=\frac{S_{d-1} }{2\cdot(\pi)^{d/2}} \cdot \Gamma(d/2, r_g^2/2) \\
&\mathrm{(general \ d)} 
\end{split}
\\
\begin{split}
\mathrm{CDF}_{r,g,1}&=\mathrm{erf}(r_g/\sqrt{2}) \\ 
&\mathrm{(d \ = \ 1)}
\end{split}
\\
\begin{split}
\mathrm{CDF}_{r,g,2}&=1-\mathrm{exp}(-r_g^2/2) \\
&\mathrm{(d \ = \ 2)}
\end{split}
\end{align}
and

\begin{align}
\begin{split}
\mathrm{CDF}_{r,f}&=\frac{S_{d-1} \cdot (2\cdot r_f)^d}{S_d \cdot d} \\
&\ \ \ \cdot {}_2 F_1(d/2;d;d/2+1;-r_f^2) \\
&\mathrm{(general \ d)}
\end{split}
\\
\begin{split}
\mathrm{CDF}_{r,f,1}&=\frac{2}{\pi} \cdot \mathrm{tan}^{-1}(r_f) \\
&\mathrm{(d \ = \ 1)} 
\end{split}
\\
\begin{split}
\mathrm{CDF}_{r,f,2}&=\frac{r_f^2}{r_f^2+1}, \\
&\mathrm{(d \ = \ 2)}
\end{split}
\end{align} where $\Gamma(x,y)$ is the upper incomplete Gamma function and ${}_2 F_1$ is the Gauss hypergeometric function.
The transformation $\rho_1$ then can be written with the corresponding CDFs as

\begin{align}
r_f=\rho_1(r_g)=\mathrm{CDF}_{r,f}^{-1}(\mathrm{CDF}_{r,g}(r_g)),
\end{align}
which can in general not be written down analytically except for $d=1$ and $d=2$. Using bisection and Newton iterations it is possible to evaluate this expression and its inverse for higher $d$. The Newton iterations make the result differentiable, which is required if it is to be used in variational autoencoders. Finally, the transformation $\rho_{\mathrm{tot}}=\rho_2 \circ \rho_1$ can be written as 

\begin{align}
\begin{split}
\theta_{d-1}=\rho_{\mathrm{tot}}(r_g)&=\rho_2\left(\mathrm{CDF}_{r,f}^{-1}(\mathrm{CDF}_{r,g}(r_g)\right) \\
&\mathrm{(general \ d)}
\end{split}
\\
\begin{split}
\rho_{\mathrm{tot},1}(r_g) &= \pi\cdot(1-\mathrm{erf}(r_g/\sqrt{2}))  \\
&\mathrm{(d \ = \ 1)}
\end{split}
\\
\begin{split}
\rho_{\mathrm{tot},2}(r_g) &= \mathrm{arccos}\left(1-2\cdot \mathrm{exp}(-r_g^2/2)\right) , \\
&\mathrm{(d \ = \ 2)} 
\end{split}
\end{align}
which has a simple and invertible structure for $d=1$ and $d=2$ after some manipulation with trigonometric identities, while higher dimensions again require bisection and Newton iterations. Because $\rho_{tot}$ defines a flow from the d-dimensional standard normal distribution to the flat distribution on the d-sphere, besides its usage in normalizing flows for coverage as describe in Section \ref{sec:coverage}, it can be used as a non-standard way to generate uniform samples on the d-sphere. This can be done by first sampling from the d-dimensional standard normal distribution, switching to spherical coordinates, and finally transforming the radial coordinate to the last coordinate $\theta_{d-1}$ on the sphere using $\rho_{tot}$, while keeping all other angles $\theta_1, \ldots, \theta_{d-2}$ and $\phi$ as they are.

\bibliography{article}

\end{document}